\documentclass[10pt,twocolumn,letterpaper]{article}

\usepackage[pagenumbers]{cvpr} % To force page numbers, e.g. for an arXiv version

\definecolor{cvprblue}{rgb}{0.21,0.49,0.74}
\usepackage[pagebackref,breaklinks,colorlinks,allcolors=cvprblue]{hyperref}
\usepackage{xcolor} % Kang Add for colored text
\usepackage{graphicx}
\usepackage{amsmath}
\usepackage{amsthm}
\usepackage{booktabs}
\usepackage{algorithm}
\usepackage{algorithmic}
\usepackage[switch]{lineno}
\usepackage[table]{xcolor}
\usepackage{multicol}
\usepackage{multirow}
\def\paperID{35474} % *** Enter the Paper ID here
\def\confName{CVPR}
\def\confYear{2026}

\title{Hg-I2P: Bridging Modalities for Generalizable Image-to-Point-Cloud Registration via Heterogeneous Graphs}

\author{
Pei An\textsuperscript{1}, 
Junfeng Ding\textsuperscript{1}, 
Jiaqi Yang\textsuperscript{2}$^*$, 
Yulong Wang\textsuperscript{3}, 
Jie Ma\textsuperscript{1}, 
Liangliang Nan\textsuperscript{4}\thanks{Corresponding authors: Jiaqi Yang (jqyang@nwpu.edu.cn) and Liangliang Nan (liangliang.nan@tudelft.nl)}\\
\textsuperscript{1}{Huazhong University of Science and Technology, China}\\
\textsuperscript{2}{Northwestern Polytechnical University, China}\\
\textsuperscript{3}{Huazhong Agricultural University, China} \quad 
\textsuperscript{4}{Delft University of Technology, Netherlands}
}

\begin{document}
\maketitle

\begin{abstract}
Image-to-point-cloud (I2P) registration aims to align 2D images with 3D point clouds by establishing reliable 2D-3D correspondences. The drastic modality gap between images and point clouds makes it challenging to learn features that are both discriminative and generalizable, leading to severe performance drops in unseen scenarios. 

We address this challenge by introducing a heterogeneous graph that enables refining both cross-modal features and correspondences within a unified architecture. The proposed graph represents a mapping between segmented 2D and 3D regions, which enhances cross-modal feature interaction and thus improves feature discriminability. In addition, modeling the consistency among vertices and edges within the graph enables pruning of unreliable correspondences. Building on these insights, we propose a heterogeneous graph embedded I2P registration method, termed Hg-I2P. It learns a heterogeneous graph by mining multi-path feature relationships, adapts features under the guidance of heterogeneous edges, and prunes correspondences using graph-based projection consistency.
Experiments on six indoor and outdoor benchmarks under cross-domain setups demonstrate that Hg-I2P significantly outperforms existing methods in both generalization and accuracy. Code is released on \url{https://github.com/anpei96/hg-i2p-demo}.
\end{abstract}
\section{Introduction}
\label{sec.intro}

% background and motivation
Image-to-point-cloud (I2P) registration aims to establish pixel-to-point correspondences between an image and a 3D point cloud~\cite{2d3d-match, 2d3d-matr}, serving as a cornerstone for tasks such as visual localization~\cite{ep2p-loc}, navigation~\cite{iplannar}, and 3D reconstruction~\cite{furao, mast3r-slam, cmr-next}. 
Recent learning-based methods approach this task by extracting features for both modalities and identifying correspondences through feature matching~\citep{2d3d-match, p2-net, 2d3d-matr, mincd}. While progress has been made through improved backbones~\cite{freereg}, matching strategies~\cite{graph-i2p-cvpr, diff-reg}, and loss functions~\cite{corr-learning-i2p-cvpr}, these methods still struggle to generalize to unseen environments. 

\begin{figure}[t]
	\centering
		\includegraphics[width=1.0\linewidth]{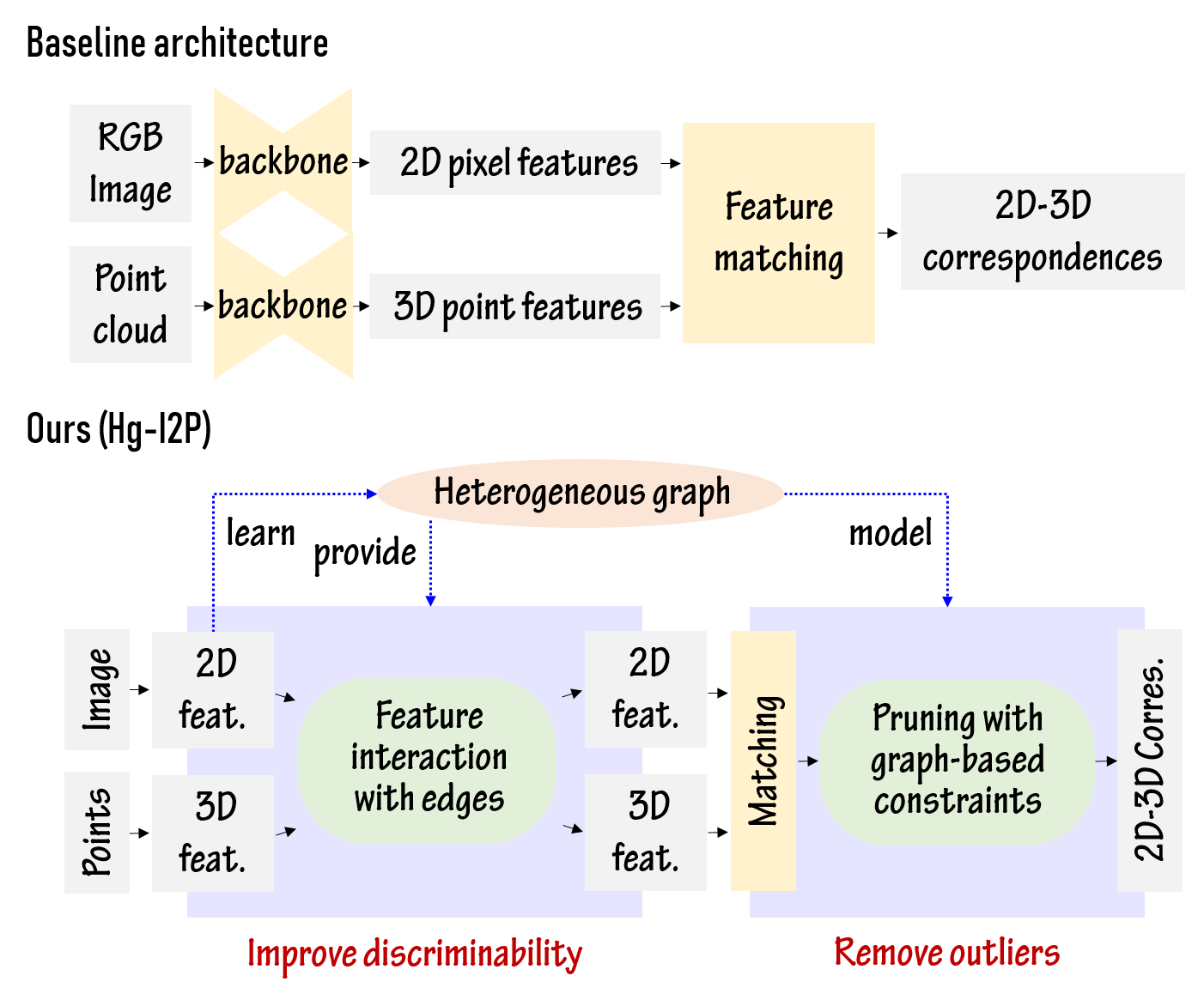}	
		\caption{
\textbf{Motivation of Hg-I2P}. To achieve generalizable I2P registration, we reformulate the baseline architecture using a heterogeneous graph.
Its heterogeneous edges enhance cross-modal feature interaction, improving feature discriminability.
Projection constraints within the graph enable consistency-based pruning of mismatched correspondences, enhancing robustness.
The resulting framework, Hg-I2P, jointly refines both features and matches, achieving strong generalization across unseen scenes.}
	\label{fig:intro}
\end{figure}

% problem and gap
The main reason for this limitation lies in the significant modality gap between images and point clouds. The appearance-driven 2D features and geometry-driven 3D features follow fundamentally different distributions. Consequently, even the correct correspondences may exhibit low feature similarity, making it difficult for neural networks to distinguish true correspondences~\cite{bridge}. Existing methods attempt to alleviate this problem either by refining cross-model features~\cite{top-i2p} or pruning incorrect correspondences~\cite{graph-i2p-cvpr}, yet each only tackles one aspect in isolation. Furthermore, feature refinement methods~\cite{top-i2p, graph-i2p-cvpr, bridge} often lack explicit cross-modal reasoning, while correspondence pruning typically relies on depth prediction or handcrafted heuristics. This fragmented treatment leaves the generalization problem largely unsolved.

% what we aim to achieve: a unified cross-modal graph
%Traditional methods treat feature learning and correspondence pruning separately. Hg-I2P unifies them through a heterogeneous graph that models region-level 2D–3D relations. This allows cross-modal feature exchange (improving discriminability) and graph-based consistency reasoning (improving robustness), yielding stronger generalization to unseen domains.

To address these issues, we argue that a unified structure capable of jointly reasoning 2D and 3D relationships is essential for robust and generalizable registration. To this end, we introduce a heterogeneous graph for I2P data. It captures a mapping between 2D and 3D regions, thereby strengthening cross-modal feature interaction and enriching representations with both geometric and visual cues. The consistency of vertices and edges within the graph provides a principled means to prune incorrect correspondences.

% our method
Building on these insights, we develop a \textbf{H}eterogeneous \textbf{g}raph embedded \textbf{I2P} registration method (\textbf{Hg-I2P}). It incorporates three core components: 
\textbf{(1) MP-mining}, which learns heterogeneous edges by mining multi-path adjacency relations on a graph; 
\textbf{(2) HG-adapting}, which refines I2P features by coarse-to-fine cross-modal message passing along heterogeneous edges; and
\textbf{(3) HC-pruning}, which enforces graph-aware projection consistency to effectively filter out incorrect correspondences.
Together, these components enable a joint refinement of both features and correspondences on the graph, significantly enhancing generalization across domains.
Our contributions are as follows:
\begin{itemize}
    \item We identify \textbf{a unified perspective for generalizable I2P} registration, where a heterogeneous graph simultaneously refines both cross-modal features and correspondences.
    \item We propose Hg-I2P, \textbf{a heterogeneous graph-embedded registration architecture} that incorporates graph learning, feature adaptation, and correspondence pruning. 
    \item We conduct \textbf{extensive cross-domain experiments} on six indoor and outdoor datasets, demonstrating strong generalization and state-of-the-art performance. 
\end{itemize}

\section{Related Works}
\label{sec.related}

We first review the architectures of mainstream learning-based I2P registration methods. Then, we discuss the representative works that aim to improve the generalization ability of I2P registration.

\vspace{+1mm}
\noindent\textbf{Architectures of learning-based I2P registration}. The design of I2P registration architectures usually draws inspiration from learning-based image or point cloud registration methods~\cite{deep-i2p, 2d3d-matr}. In 2019, Feng et al.~\cite{2d3d-match} learned a modality-invariant descriptor for both 2D and 3D pre-detected keypoints. But, pre-detected keypoints often contain few inliers and negatively affect learning-based I2P registration performance~\cite{deep-i2p}. To avoid this issue, researchers began to jointly learn both keypoints and descriptors, leading to progress in I2P architecture design. Wang et al.~\cite{p2-net} introduced a mechanism to learn keypoints by incorporating spatial-wise and channel-wise maxima on I2P feature maps. Li et al.~\cite{deep-i2p} and Ren et al.~\cite{corr-tcsvt} proposed cross-modal global attention modules for more discriminative I2P feature learning. Building on this, Zhou et al.~\cite{diff_reg_match} refined loss functions by embedding adaptive-weight optimization into the circle loss~\cite{circle-loss}, dynamically adjusting correspondence weights. They also incorporated differentiable perspective-n-point (PnP)~\cite{diff-pnp-layer} to further suppress outliers. Some researchers designed I2P registration models inspired by point cloud registration architectures~\cite{geotrans}, such as MATR~\cite{2d3d-matr}, CoFiI2P~\cite{CoFiI2P}, and PAPI-Reg~\cite{PAPI-Reg}. These methods predict 2D-3D correspondences in a coarse-to-fine manner (i.e., first matching 2D-3D patches, then generating correspondences guided by the matched regions), thereby improving accuracy on public benchmarks.

\vspace{+1mm}
\noindent\textbf{Improving generalization in I2P registration}. 
As more researchers have observed unstable I2P performance in unseen environments~\cite{freereg, ol-reg, bridge, top-i2p}, recent works have focused on improving feature generalization across domains. Vision foundation models, such as DepthAnything~\cite{depthanything}, Segment Anything Model (SAM)~\cite{sam_2d}, and object detectors~\cite{centerpts} are increasingly leveraged to bridge modality gaps in I2P data. For example, Wang et al.~\cite{freereg} used a pre-trained ControlNet~\cite{controlnet} and depth prediction model to realign cross-modal features, though their approach requires over $\geq 10$ seconds per image. An et al.~\cite{ol-reg} employed 3D object detectors to align objects between images and point clouds, designing Proj-I2P to extract correspondences from aligned object pairs. Wu et al.~\cite{diff-reg} utilized the DINO model~\cite{dino} to obtain spatially-aware RGB features, while Bie et al.~\cite{graph-i2p-cvpr} leveraged DepthAnything~\cite{depthanything} to convert I2P registration into a 3D-3D point cloud registration task. Because the predicted depths lack real-world scale, they proposed a distribution alignment module to minimize the difference between LiDAR and back-projected point clouds. Murai et al.~\cite{mast3r-slam} applied the multi-view reconstruction model MASt3R~\cite{mast3r} to generate point clouds from images and then used a gradient-based ICP scheme in $\textbf{Sim}(3)$ space to estimate camera poses. An et al.~\cite{top-i2p} employed both the 2D and 3D SAM models~\cite{sam_2d, sam_3d} to segment images and point clouds into regions. They predict topological relationships between these segments to assist cross-modal alignment. 

Some works improve I2P feature generalization without relying on vision foundation models. Cheng et al.~\cite{bridge} modeled the matching uncertainty between 2D and 3D patches. Yao et al.~\cite{cmr-agent} formulated correspondence generation as a Markov decision process and developed a reinforcement learning based agent for I2P registration. Li et al.~\cite{corr-learning-i2p-cvpr} proposed an implicit correspondence learning module to predict 2D and 3D keypoints jointly.

\vspace{+1mm}
\noindent\textbf{Discussions}. To improve I2P registration generalization, leveraging vision foundation models has become a mainstream strategy, as these models possess strong generalization capabilities in unseen domains~\cite{diffi2p}. In this paper, we utilize 2D and 3D SAM models to construct a heterogeneous graph as a bridge for cross-modal feature matching. Compared with the prior work~\cite{top-i2p} that also employed SAM, our method has three key advantages: 
(1) We define a heterogeneous graph that systematically models relationships between 2D and 3D regions; 
(2) We exploit both edge information and projection constraints in this graph to refine both I2P features and correspondences; 
(3) Hg-I2P achieves stronger generalization and higher accuracy than state-of-the-art methods~\cite{top-i2p, 2d3d-matr, mincd, graph-i2p-cvpr, bridge} across multiple datasets. 

\section{Heterogeneous Graph Representation}
\label{sec.graph}

\begin{figure}[t]
	\centering
		\includegraphics[width=1.0\linewidth]{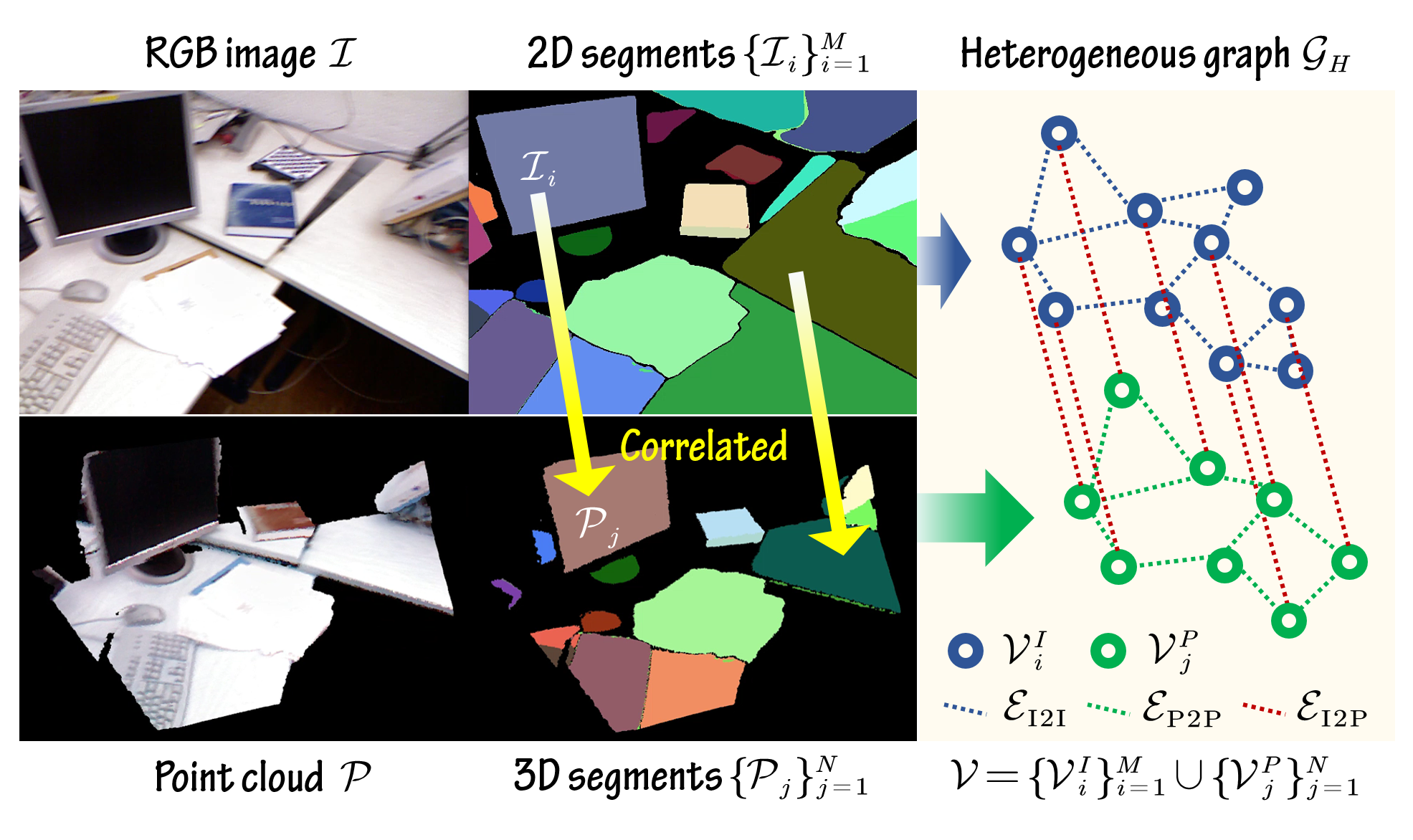}	
		\caption{Constructing a heterogeneous graph ($\mathcal{G}_H$) for cross-modal reasoning. This graph models the relationship between 2D and 3D regions. By explicitly linking visual and geometric entities through heterogeneous edges, it captures structured 2D-3D dependencies essential for joint reasoning. }
	\label{fig:evidence}
\end{figure}
% It serves as a coarse mapping between the image and the point cloud.
% The strong shape correlations of 2D and 3D segments strengthen the construction accuracy of $\mathcal{G}_H$

Before illustrating the effect of the heterogeneous graph on I2P registration, we define this structure formally. Given an RGB image and a colored point cloud, we use 2D and 3D SAM~\cite{sam_2d, sam_3d, sam3d_used} to decompose the image and point cloud into multiple 2D and 3D regions, and then define a heterogeneous weighted 
graph $\mathcal{G}_H=(\mathcal{V}_H, \mathcal{E}_H)$ to systematically model relationships between these regions (see Fig.~\ref{fig:evidence}). %More discussion is provided in the supplemental material.

\vspace{+1mm}
\noindent\textbf{Definition}. In $\mathcal{G}_H$, the vertices and edges  are defined as $\mathcal{V}_H=\mathcal{V}_I \cup \mathcal{V}_P=\{\mathcal{V}^I_i\}_{i=1}^M \cup \{\mathcal{V}^P_j\}_{j=1}^N$ and $\mathcal{E}_H = \mathcal{E}_{\text{I2I}} \cup \mathcal{E}_{\text{P2P}} \cup \mathcal{E}_{\text{I2P}}$. We assign each vertex $\mathcal{V}^I_i$ (or $\mathcal{V}^P_j$) with a $c$-dimensional feature vector $\mathbf{v}^I_i \in \mathbb{R}^c$ (or $\mathbf{v}^P_j$). In practice, we use 2D and 3D SAM~\cite{sam_2d, sam_3d} to segment $\mathcal{I}$ and $\mathcal{P}$ into non-overlapping regions $\{\mathcal{I}_i\}_{i=1}^M$ and $\{\mathcal{P}_j\}_{j=1}^N$:
\begin{equation}
\label{eq_1}
\mathcal{I} = \cup_{i}^M \mathcal{I}_i,\,\, \mathcal{P} = \cup_{j}^N \mathcal{P}_j, 
\end{equation}
\noindent where each region $\mathcal{I}_i$ (or $\mathcal{P}_j$) corresponds to a vertex $\mathcal{V}^I_i$ (or $\mathcal{V}^P_j$). In $\mathcal{G}_H$, $\mathcal{E}_{\text{I2I}}$ (or $\mathcal{E}_{\text{P2P}}$) are \textbf{homogeneous} edge set describing intra-modal relationship between $\mathcal{I}_i$ and $\mathcal{I}_j$ (or $\mathcal{P}_i$ and $\mathcal{P}_j$). $\mathcal{E}_{\text{I2P}}$ is the \textbf{heterogeneous} edge set describing cross-modal relationships between $\{\mathcal{I}_i\}_{i=1}^M$ and $\{\mathcal{P}_j\}_{j=1}^N$.  $\mathcal{E}_{\text{I2I}}$, $\mathcal{E}_{\text{P2P}}$, and $\mathcal{E}_{\text{I2P}}$ are represented by three adjacent matrices $\mathbf{E}_{\text{I2I}}\in \mathbb{R}^{M\times M}$, $\mathbf{E}_{\text{P2P}}\in \mathbb{R}^{N\times N}$, and $\mathbf{E}_{\text{I2P}} \in \mathbb{R}^{M\times N}$, respectively. As $\mathcal{G}_H$ is a weighted graph, the homogeneous edges can be defined based on feature distance:
\begin{equation}
\label{eq_2}
(\mathbf{E}_{\text{I2I}})_{ij} = e^{-\alpha\Vert \mathbf{v}^I_i - \mathbf{v}^I_j \Vert_2^2},
(\mathbf{E}_{\text{P2P}})_{ij} = e^{-\alpha\Vert \mathbf{v}^P_i - \mathbf{v}^P_j \Vert_2^2}
\end{equation}
\noindent where $\alpha$ is a distance scaling factor, and $(\cdot)_{ij}$ queries a value from a matrix at index $(i,j)$.

\vspace{+1mm}
\noindent\textbf{Heterogeneous edges}.  In general, $\mathcal{E}_{\text{I2P}}$ reflects the correlation of 2D-3D regions. Extending from work~\cite{top-i2p}, we model this correlation through the overlap between a 2D region $\mathcal{I}_i$ and the projection of a 3D region $\mathcal{V}_j$, quantified using the interaction-over-union (IoU). Thus, 
\begin{equation}
\label{eq_3}
(\mathbf{E}_{\text{I2P}})_{ij} = \mathrm{IoU}(\mathcal{I}_i, \pi(\mathcal{P}_j|\mathbf{T}_{gt}))
\end{equation}
\noindent where $\pi(\cdot|\mathbf{T})$ denotes the camera projection under a pose $\mathbf{T}\in \mathbf{SE}(3)$~\cite{camera_model}, and  $\mathbf{T}_{gt}$ is the ground truth (GT) pose. If no overlap exists (i.e., $\mathcal{I}_i \cap \pi(\mathcal{P}_j|\mathbf{T}_{gt}) = \emptyset$), $(\mathbf{E}_{\text{I2P}})_{ij}=0$, making $\mathbf{E}_{\text{I2P}}$ a sparse matrix. This construction provides a \textbf{coarse mapping between 2D and 3D regions}, enabling the refinement of both I2P features and correspondences.

\begin{figure*}[t]
	\centering
		\includegraphics[width=1.0\linewidth]{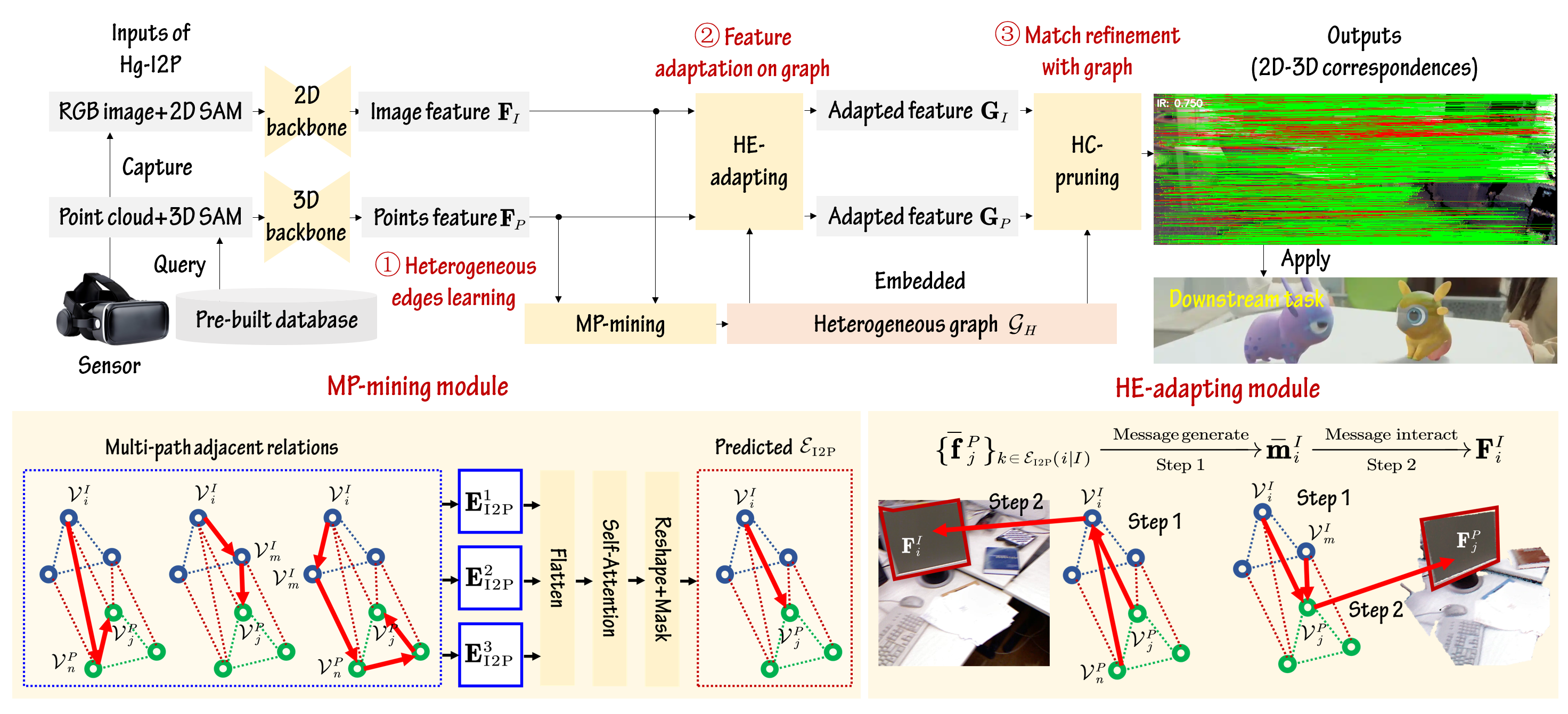}	
		\caption{\textbf{Pipeline of Hg-I2P}. MP-mining learns heterogeneous edges $\mathcal{E}_{\text{I2P}}$ and constructs the graph $\mathcal{G}_H$. HE-adapting refines cross-modal features via message passing along heterogeneous edges, improving cross-modal alignment.
        HC-pruning enforces projection consistency within $\mathcal{G}_H$ to remove outliers. Together, these components refine both features and correspondences for robust I2P registration.}
	\label{fig:method}
\end{figure*}

\vspace{+1mm}
\noindent\textbf{Initialization}. Using 2D, 3D backbone networks~\cite{kpconv, resnet}, we extract feature maps $\mathbf{F}_I \in \mathbb{R}^{H\times W\times c}$ and $\mathbf{F}_P \in \mathbb{R}^{A\times c}$ from the image $\mathcal{I}$ and point cloud $\mathcal{P}$. Here, $H$ and $W$ represent the image height and width, respectively. $A$ is the points number, and $c$ is feature dimension. In practice, we set the vertex feature $\mathbf{v}^I_i$ (or $\mathbf{v}^P_j$) as $\bar{\mathbf{f}}^I_{i}$ (or $\bar{\mathbf{f}}^P_{j}$), which is average pooled from $\mathbf{F}_I$ (or $\mathbf{F}_P$) on region $\mathcal{I}_i$ (or $\mathcal{P}_j$). More detail is shown in the supplemental materials. Since $\mathbf{T}_{gt}$ is unavailable in I2P registration, $\mathcal{E}_{\text{I2P}}$ cannot be directly computed. We learn $\mathcal{E}_{\text{I2P}}$ in the Hg-I2P architecture.

%In all, $\mathcal{G}_H$ models sparse region-level 2D-3D correspondences. It serves as a crucial cue for I2P feature adaptation and match refinement. However, as $\mathbf{T}_{gt}$ is unknown in I2P registration, $\mathcal{E}_{\text{I2P}}$ cannot be directly constructed. In Sec. 4, we design an effective scheme for $\mathcal{G}_H$ construction. 

\section{Proposed Method}
\label{sec.method}

\subsection{Overview of Hg-I2P}
\label{sec.method.A}
%Given an image $\mathcal{I}$, a point cloud $\mathcal{P}$ and pose prior $\mathbf{T}_0$, I2P registration is to establish 2D-3D correspondences. 

% multi-path correlation mining (MP-mining)
% heterogeneous edge guided adaptation (HE-adapting)
% heterogeneous edge-aware criteria (HC-pruning)

To effectively utilize the heterogeneous graph for generalized I2P registration, we propose a complete pipeline called Hg-I2P. As shown in Fig.~\ref{fig:method}, it consists of three core components: (1) \textbf{Heterogeneous edges learning}, which predicts $\mathcal{E}_{\text{I2P}}$ and construct $\mathcal{G}_H$; (2) \textbf{Feature adaptation on graph}, which refines the cross-modal features under the guidance of $\mathcal{E}_{\text{I2P}}$; (3) \textbf{Match refinement with graph}, which enforces projection consistency in $\mathcal{G}_H$ to prune outliers. Overall, by leveraging $\mathcal{G}_H$, the proposed feature adaptation and match refinement steps enhance both feature discriminability and correspondence accuracy, thereby improving the generalization capability of I2P registration.

\subsection{Learning heterogeneous edges with multi-path adjacent relations mining}
\label{sec.method.B}

To construct $\mathcal{G}_H$, accurately learning $\mathcal{E}_{\text{I2P}}$ from I2P data is a crucial step. To fully exploit $\mathcal{E}_{\text{I2I}}$ and $\mathcal{E}_{\text{P2P}}$ for this purpose, we \textbf{mine} \textbf{m}ulti-\textbf{p}ath (MP) adjacent relations and introduce an \textbf{MP-mining} module to predict $\mathcal{E}_{\text{I2I}}$. 

To incorporate $\mathcal{E}_{\text{I2I}}$ and $\mathcal{E}_{\text{P2P}}$, we model the adjacency between  $\langle \mathcal{I}_i, \mathcal{P}_j \rangle$ by traversing multiple paths from $\mathcal{V}^I_i$ to $\mathcal{V}^P_j$. For example, path $\mathcal{V}^I_i \rightarrow \mathcal{V}^I_m \rightarrow \mathcal{V}^P_j$ is one feasible route on $\mathcal{G}_H$ $(m=1,...,M)$. In this case, the adjacency relation is modeled as $(\mathbf{E}_{\text{I2P}})_{ij} \leftarrow \sum_{m=1}^M (\mathbf{E}_{\text{I2I}})_{im} (\mathbf{E}_{\text{I2P}})_{mj}$. Similarly, other paths such as $\mathcal{V}^I_i \rightarrow \mathcal{V}^P_n \rightarrow \mathcal{V}^P_j$ and $\mathcal{V}^I_i \rightarrow \mathcal{V}^I_m \rightarrow \mathcal{V}^P_n \rightarrow \mathcal{V}^P_j$ are also considered. 
We initialize the heterogeneous edges as $\mathbf{\tilde{E}}_{\text{I2P}}$, where $(\mathbf{\tilde{E}}_{\text{I2P}})_{ij} = e^{-\alpha\Vert \mathbf{v}^I_i - \mathbf{v}^P_j \Vert_2^2}$ and model the adjacency relations as three matrices: $\mathbf{E}_{\text{I2P}}^1 = \mathbf{E}_{\text{I2I}} \mathbf{\tilde{E}}_{\text{I2P}}$, $\mathbf{E}_{\text{I2P}}^2 = \mathbf{\tilde{E}}_{\text{I2P}} \mathbf{E}_{\text{P2P}}$,  $\mathbf{E}_{\text{I2P}}^3 = \mathbf{E}_{\text{I2I}} \mathbf{\tilde{E}}_{\text{I2P}}  \mathbf{E}_{\text{P2P}}$. 

From a \textbf{Bayesian inference} perspective\footnote{As $(\mathbf{E}_{\text{I2I}})_{ij}\in (0,1)$, it can be regarded as a conditional probability $P(\mathcal{V}^I_i|\mathcal{V}^P_j)$. This way, $\mathbf{E}_{\text{I2P}}^3$ (as an example) can be explained as: \\$(\mathbf{E}_{\text{I2P}}^3)_{ij}$ is $P(\mathcal{V}^I_i|\mathcal{V}^P_j) = \sum_{m} \sum_{n} P(\mathcal{V}^I_i|\mathcal{V}^I_m)  P(\mathcal{V}^I_m|\mathcal{V}^P_n)  P(\mathcal{V}^P_n|\mathcal{V}^P_j)$.}, $\{\mathbf{E}_{\text{I2P}}^k\}_{k=1}^3$ capture the significant casual relations between $\mathcal{I}_i$ and $\mathcal{P}_j$. Based on this insight, MP-mining predicts $\mathbf{E}_{\text{I2P}}$ by mining correlation features from these matrices:
\begin{equation}
\label{eq_4}
\begin{aligned}
\mathbf{\hat{E}}_{\text{I2P}} &= \mathrm{RS}(\mathrm{Attention}(\mathbf{F}_{\text{I2P}}\mathbf{W}^Q_{\mathrm{MP}}, \mathbf{F}_{\text{I2P}}\mathbf{W}^K_{\mathrm{MP}}, \mathbf{F}_{\text{I2P}}\mathbf{W}^V_{\mathrm{MP}})) \\
\mathbf{F}_{\text{I2P}} &= \mathbf{W}_{\mathrm{MP}}^F \cdot \mathrm{Flatten}([\mathbf{E}_{\text{I2P}}^1 , \mathbf{E}_{\text{I2P}}^2 , \mathbf{E}_{\text{I2P}}^3 ])  
\end{aligned}
\end{equation}
\begin{equation}
\label{eq_5}
\mathrm{Attention}(\mathbf{Q},\mathbf{K},\mathbf{V}) = \mathrm{Softmax}(\mathbf{Q}\mathbf{K}^T/\sqrt{d_k})\mathbf{V}
\end{equation}
\noindent where $\mathrm{Attention}(\mathbf{Q},\mathbf{K},\mathbf{V})$ is an attention layer~\cite{transformer} and $\mathbf{Q}$, $\mathbf{K}$, $\mathbf{V} \in \mathbb{R}^{d_k}$ denote the query, key, and value vectors, respectively. $\mathrm{RS}(\cdot)$ is the reshape operation that converts an $M \times N$-dimensional vector into an $M\times N$ matrix. $\mathbf{W}_{\mathrm{MP}}^F \in \mathbb{R}^{3MN\times MN}$, $\mathbf{W}^Q_{\mathrm{MP}}$, $\mathbf{W}^K_{\mathrm{MP}}$, and $\mathbf{W}^V_{\mathrm{MP}} \in \mathbb{R}^{MN\times MN}$ are the learnable matrices. $\mathrm{Flatten}(\cdot)$ is an operation to flatten a tensor as vector. In practice, $\mathbf{\hat{E}}_{\text{I2P}}$ is $L1$-normalized row-wise before used. Together with $\mathcal{E}_{\text{I2I}}$, $\mathcal{E}_{\text{P2P}}$, and the learned $\mathcal{E}_{\text{I2P}}$, the final heterogeneous graph $\mathcal{G}_H$ can be constructed. 

% The low-confidence predictions in $\mathbf{\hat{E}}_{\text{I2P}}$ are removed and the more details refer to the supplemental materials.

%Then, to enforce the edge sparsity, we add a mask $\mathbf{M}_{\text{I2P}} \in \mathbb{R}^{M\times N}$ on $\mathbf{\hat{E}}_{\text{I2P}}$ to remove the incorrect edges with a reprojection threshold $\tau_{2d}$. $\mathbf{M}_{\text{I2P}}$ is constructed as:

%\begin{equation}
%(\mathbf{M}_{\text{I2P}})_{ij} = \mathbf{1}(\Vert \bar{\mathbf{p}}^I_i - \bar{\mathbf{p}}^P_j \Vert_2 \leq \tau_{2d})
%\end{equation}

%\noindent where $\mathbf{1}(\cdot)$ is an indicator function. $\odot$ denotes an element-wise product. After learning $\mathcal{E}_{\text{I2P}}$, we construct $\mathcal{G}_H$ finally. 

\subsection{Heterogeneous edges guided feature adaptation}
\label{sec.method.C}

As discussed in Sec.~\ref{sec.graph}, $\mathcal{E}_{\text{I2P}}$ defines a mapping between 2D and 3D segmented regions. To leverage this mapping for cross-modal feature learning, we propose a \textbf{heterogeneous edge} guided \textbf{adaptation} module, termed \textbf{HE-adapting}. 

Given $\{\mathcal{I}_i\}_{i=1}^M$ and $\{\mathcal{P}_j\}_{j=1}^N$, we decompose $\mathbf{F}_I$ and $\mathbf{F}_P$ into feature groups $\{\mathbf{F}_i^I\}_{i=1}^M$ and $\{\mathbf{F}_j^P\}_{j=1}^N$\footnote{$\mathbf{F}_i^I \in \mathbb{R}^{m_i\times c}$ and $\mathbf{F}_j^P \in \mathbb{R}^{n_j\times c}$. The $m_i$ (or $n_j$) denotes the number of pixels (or points) in the region $\mathcal{I}_i$ (or $\mathcal{P}_j$).}. Using $\mathcal{E}_{\text{I2P}}$, region $\mathcal{I}_i$ (or $\mathcal{P}_j$) is associated with its neighboring regions $\{\mathcal{P}_k\}_{k\in \mathcal{E}_{\text{I2P}}(i|I)}$ (or $\{\mathcal{I}_k\}_{k\in \mathcal{E}_{\text{I2P}}(j|P)}$). We adopt a lightweight \textit{message generation-interaction} scheme for feature refinement:
$$
\begin{aligned}
\{\bar{\mathbf{f}}_{k}^P\}_{k\in \mathcal{E}_{\text{I2P}}(i|I)} &\xrightarrow[\mathrm{Step\, 1}]{\mathrm{Message\,generate}} \bar{\mathbf{m}}_{i}^I \xrightarrow[\mathrm{Step\, 2}]{\mathrm{Message\,interact}} \mathbf{F}_i^I \\
\{\bar{\mathbf{f}}_{k}^I\}_{k\in \mathcal{E}_{\text{I2P}}(j|P)} &\xrightarrow[\mathrm{Step\, 1}]{\mathrm{Message\,generate}} \bar{\mathbf{m}}_{j}^P \xrightarrow[\mathrm{Step\, 2}]{\mathrm{Message\,interact}} \mathbf{F}_j^P 
\end{aligned}
$$
In this process, cross-modal message feature $\bar{\mathbf{m}}_{i}^I$ (or $\bar{\mathbf{m}}_{j}^P$) is first derived from vertex features $\{\bar{\mathbf{f}}_{k}^P\}_{k\in \mathcal{E}_{\text{I2P}}(i|I)}$ (or $\{\bar{\mathbf{f}}_{k}^I\}_{k\in \mathcal{E}_{\text{I2P}}(j|P)}$). Subsequently, each region feature $\mathbf{F}_i^I$ (or $\mathbf{F}_j^P$) is refined through interaction with the corresponding message feature $\bar{\mathbf{m}}_{i}^I$ (or $\bar{\mathbf{m}}_{j}^P$). The procedure is detailed below: 

\vspace{+1mm}
\noindent\textbf{Step 1: Message generation}. 
The message $\bar{\mathbf{m}}_{i}^I$ (or $\bar{\mathbf{m}}_{j}^P$) is encoded from $\{\bar{\mathbf{f}}_{j}^P\}_{k\in \mathcal{E}_{\text{I2P}}(i|I)}$ (or $\{\bar{\mathbf{f}}_{i}^I\}_{k\in \mathcal{E}_{\text{I2P}}(j|P)}$):
\begin{equation}
\label{eq_6}
\begin{aligned}
\bar{\mathbf{m}}_{i}^I &= \mathrm{Attention}(\bar{\mathbf{f}}_{i}^I \mathbf{W}^I_v, \bar{\mathbf{g}}_{i}^P \mathbf{W}^P_v, \bar{\mathbf{g}}_{i}^P \mathbf{W}^P_v) \in \mathbb{R}^{c}, \\
\bar{\mathbf{m}}_{j}^P &= \mathrm{Attention}(\bar{\mathbf{f}}_{j}^P \mathbf{W}^P_v, \bar{\mathbf{g}}_{j}^I \mathbf{W}^I_v, \bar{\mathbf{g}}_{j}^I \mathbf{W}^I_v) \in \mathbb{R}^{c}
\end{aligned}
\end{equation}
\begin{equation}
\label{eq_7}
\bar{\mathbf{g}}_{i}^P = \frac{\sum_{j=1}^N (\mathbf{\hat{E}}_{\text{I2P}})_{ij}  \bar{\mathbf{f}}_{j}^P}{\sum_{j=1}^N (\mathbf{\hat{E}}_{\text{I2P}})_{ij}}, \,\,
\bar{\mathbf{g}}_{j}^I = \frac{\sum_{i=1}^M (\mathbf{\hat{E}}_{\text{I2P}})_{ij}  \bar{\mathbf{f}}_{i}^I}{\sum_{i=1}^M (\mathbf{\hat{E}}_{\text{I2P}})_{ij}}
\end{equation}
\noindent where $\bar{\mathbf{g}}_{i}^P$ (or $\bar{\mathbf{g}}_{j}^I$) denotes the average-pooled feature from $\mathcal{V}_P$ (or $\mathcal{V}_I$) that shares edges with $\mathcal{V}_i^I$ (or $\mathcal{V}_j^P$). Through the cross-attention in Eq.~\ref{eq_6}, the message feature $\bar{\mathbf{m}}_{i}^I$ is obtained by learning the correlation between $\bar{\mathbf{g}}_{i}^P$ and $\bar{\mathbf{f}}_{i}^I$. In Eq.~\ref{eq_6}, $\mathbf{W}^I_v$ and $\mathbf{W}^P_v \in \mathbb{R}^{c\times c}$ are learnable matrices. 

\vspace{+1mm}
\noindent\textbf{Step 2: Message interaction}. The encoded messages $\{\bar{\mathbf{m}}_{i}^I\}_{i=1}^M$ (or $\{\bar{\mathbf{m}}_{j}^P\}_{j=1}^N$) are used to adapt the region-wise features $\{\mathbf{F}_i^I\}_{i=1}^M$ (or $\{\mathbf{F}_j^P\}_{j=1}^N$):
\begin{equation}
\label{eq_8}
\begin{aligned}
\mathbf{G}_{i}^I &= (1-\beta) \mathbf{F}_{i}^I \\
& + \beta \cdot \mathrm{Attention}(\mathbf{H}_{i}^I \mathbf{W}^Q_I, \mathbf{H}_{i}^I \mathbf{W}^K_I, \mathbf{H}_{i}^I \mathbf{W}^V_I), \\
\mathbf{G}_{j}^P &= (1-\beta) \mathbf{F}_{j}^P \\
& + \beta \cdot \mathrm{Attention}(\mathbf{H}_{j}^P \mathbf{W}^Q_P, \mathbf{H}_{j}^P \mathbf{W}^K_P, \mathbf{H}_{j}^P \mathbf{W}^V_P)
\end{aligned}
\end{equation}
\begin{equation}
\label{eq_9}
\begin{aligned}
\mathbf{H}_{i}^I &= \mathrm{Concat}(\mathbf{F}_{i}^I, \mathrm{Repeat}(\bar{\mathbf{m}}_{i}^I)) \in \mathbb{R}^{m_i\times 2c},   \\
\mathbf{H}_{j}^P &= \mathrm{Concat}(\mathbf{F}_{j}^P, \mathrm{Repeat}(\bar{\mathbf{m}}_{j}^P)) \in \mathbb{R}^{n_j\times 2c}
\end{aligned}
\end{equation}
\noindent where $\beta \in [0,1]$ is a hyperparameter balancing the original and adapted features. $\mathbf{W}_{I}^Q$, $\mathbf{W}_{I}^K$, $\mathbf{W}_{I}^V$, $\mathbf{W}_{P}^Q$, $\mathbf{W}_{P}^K$, and $\mathbf{W}_{P}^V \in \mathbb{R}^{2c\times c}$ are learnable matrices. $\mathrm{Repeat}(\cdot)$ replicates a tensor. $\mathrm{Concat}(\cdot, \cdot)$ concatenates feature vectors along the channel dimension. Through the self-attention in Eq.~\ref{eq_8}, HE-adapting captures the relations between each pixel (or point) in $\mathcal{I}_i$ (or $\mathcal{P}_j$) and its corresponding message feature $\bar{\mathbf{m}}_{i}^I$ (or $\bar{\mathbf{m}}_{j}^P$), yielding the adapted features $\mathbf{G}_{i}^I$ and $\mathbf{G}_{j}^P$. The resulting features $\{\mathbf{G}_{i}^I\}_{i=1}^M$ and $\{\mathbf{G}_{j}^P\}_{j=1}^N$ are concatenated to form the refined representations $\mathbf{G}_I \in \mathbb{R}^{H\times W\times c}$ and $\mathbf{G}_P \in \mathbb{R}^{A\times c}$, respectively.

Finally, the refined feature sets $\mathbf{G}_I$ and $\mathbf{G}_P$ are utilized for feature-level matching~\cite{2d3d-matr} to produce 2D-3D correspondences $\{\langle \mathbf{p}_i^c, \mathbf{P}_i^c \rangle\}_{i=1}^C$, where $C$ denotes the number of matches. 

\begin{figure}[t]
	\centering
		\includegraphics[width=1.0\linewidth]{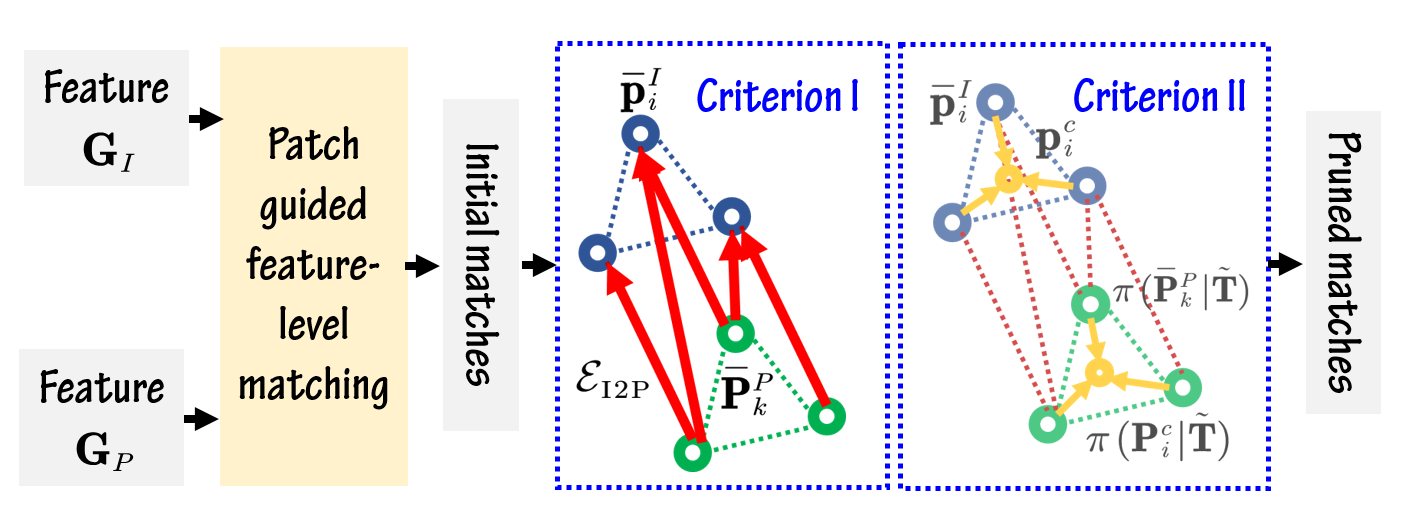}	
		\caption{Overview of HC-pruning. It filters incorrect correspondences using two projection consistency criteria derived from $\mathcal{G}_H$.}
	\label{fig:hc-pruning}
\end{figure}

\subsection{Match refinement with heterogeneous graph\\based pruning criteria}
\label{sec.method.D}

Removing outliers from $\{\langle \mathbf{p}_i^c, \mathbf{P}_i^c \rangle\}_{i=1}^C$ is crucial for robust I2P registration. To achieve this, we propose a non-learning module called \textbf{HC-pruning}, which refines correspondences based on \textbf{heterogeneous} graph-based \textbf{pruning criteria}, as shown in Fig.~\ref{fig:hc-pruning}. 

In HC-pruning, an initial camera pose $\mathbf{\tilde{T}}\in \mathbf{SE}(3)$ is estimated 
via a RANSAC-based perspective-n-point (PnP) algorithm~\cite{epnp}, which leverages the strong shape correlation between 2D and 3D segments \cite{top-i2p}. Then, with $\mathbf{\tilde{T}}$, two pruning criteria are applied: (1) based on the adjacency in $\mathcal{E}_{\text{I2P}}$ and reprojection distance $\delta_{\text{rej}}$, and (2) based on the cosine similarity between relative position vectors derived from graph-based projections.
A correspondence is identified as an inlier if at least one of the criteria is satisfied. 
The details of these criteria are elaborated in the supplementary materials. This dual-criterion approach effectively removes false matches caused by pose estimation noise or imperfect edge learning.

\subsection{Loss functions}
\label{sec.method.E}

To train Hg-I2P, we define the overall loss as:
\begin{equation}
\label{eq_11}
L_{\text{Hg-I2P}} = L_{\mathrm{corr}} + \lambda_1\Vert \mathbf{\hat{E}}_{\text{I2P}}[\mathrm{mask}]-\mathbf{{E}}_{\text{I2P}}[\mathrm{mask}]\Vert_2^2 %+ \lambda_2 L_{\mathrm{pose}}(\mathbf{T}_{gt})
\end{equation}
\noindent where $L_{\mathrm{corr}}$ is the standard correspondence loss used in I2P registration~\cite{2d3d-matr}, which contains the circle loss~\cite{circle-loss} to penalize the correspondences generated from $\mathbf{G}_I$ and $\mathbf{G}_P$ by a patch guided feature-level matching. The second term supervises the learning of $\mathcal{E}_{\text{I2P}}$. The $\mathrm{mask}$ denotes the indices of valid (non-zero) elements in $\mathbf{{E}}_{\text{I2P}}$, and $\lambda_1$ is a weighting factor controlling edge supervision strength. This formulation ensures that both cross-modal alignment and edge prediction are jointly optimized, leading to consistent improvements in generalization and registration robustness.

%$\Vert \mathbf{\hat{E}}_{\text{I2P}}-\mathbf{{E}}_{\text{I2P}}\Vert_2^2$ is used to supervise the learning of $\mathcal{E}_{\text{I2P}}$. $L_{\mathrm{pose}}(\mathbf{T}_{gt})$ is also used to correct $\mathcal{E}_{\text{I2P}}$ by minimizing the reprojection error. $\lambda_1$ and $\lambda_2$ are the loss weight parameters. 

\begin{table}[t]
%\scriptsize
\centering
\caption{Cross-scene I2P registration results on the 7-Scenes dataset~\cite{scene-7-dataset}. $\dag$ denotes the Hg-I2P variant without HC-pruning.}
\resizebox{0.95\linewidth}{!}{%}
\begin{tabular}{c|cccccc|c}
%\toprule
\hline
IR & C$\rightarrow$F & C$\rightarrow$H & C$\rightarrow$O & C$\rightarrow$P & C$\rightarrow$K &  C$\rightarrow$S & AVG \\
\hline
%P2-Net     & 0.436 & 0.330 & 0.414 & 0.421 & 0.405 & 0.251 & 0.376 \\
MATR  & 0.455 & 0.359 & 0.420 & 0.411 & 0.390 & 0.288 & 0.387 \\
Top-I2P  & {{0.491}} & {{0.427}} & {{0.455}} & {{0.461}} & {{0.437}} & {{0.327}} & {{0.433}} \\
MinCD   & 0.542 & 0.424 & 0.502 & {{0.408}} & 0.416 & 0.379 & 0.445 \\
Hg-I2P$^\dag$  & {{0.547}} & {{0.467}} & {{0.492}} & {{0.520}} & {{0.475}} & {{0.332}} & {{0.472}} \\
\rowcolor{gray!20} Hg-I2P (Ours) & \textcolor{blue}{\textbf{0.592}} & \textcolor{blue}{\textbf{0.588}} & \textcolor{blue}{\textbf{0.744}} & \textcolor{blue}{\textbf{0.533}} & \textcolor{blue}{\textbf{0.557}} & \textcolor{blue}{\textbf{0.469}} & \textcolor{blue}{\textbf{0.581}} \\
\hline
RR & C$\rightarrow$F & C$\rightarrow$H & C$\rightarrow$O & C$\rightarrow$P & C$\rightarrow$K &  C$\rightarrow$S & AVG \\
\hline
%P2-Net    & 0.536 & 0.162 & 0.672 & 0.561 & 0.563 & 0.293 & 0.464 \\
MATR  & 0.537 & 0.167 & 0.759 & 0.581 & 0.612 & 0.214 & 0.478 \\
Top-I2P  & {{0.611}} & {{0.583}} & {{0.843}} & {{0.558}} & \textcolor{blue}{\textbf{0.678}} & {{0.500}} & {{0.628}} \\
MinCD    & {{0.671}} & {{0.250}} & {{0.869}} & {{0.574}} & {{0.619}} & {{0.571}} & {{0.592}} \\
Hg-I2P$^\dag$  & {{0.712}} & \textcolor{blue}{\textbf{0.585}} & {{0.717}} & {{0.617}} & {{0.655}} & \textcolor{blue}{\textbf{0.572}} & {{0.642}} \\
\rowcolor{gray!20} Hg-I2P (Ours) & \textcolor{blue}{\textbf{0.726}} & {{0.500}} & \textcolor{blue}{\textbf{0.949}} & \textcolor{blue}{\textbf{0.681}} & {{0.631}} & {{0.512}} & \textcolor{blue}{\textbf{0.667}} \\
%\hline
\hline
IR & O$\rightarrow$C & O$\rightarrow$F & O$\rightarrow$H & O$\rightarrow$P & O$\rightarrow$K &  O$\rightarrow$S & AVG \\
\hline
%P2-Net     & 0.416 & 0.413 & 0.403 & 0.434 & 0.386 & 0.308 & 0.393 \\
MATR  & 0.498 & 0.491 & 0.521 & 0.442 & 0.448 & 0.338 & 0.456 \\
Top-I2P  & {{0.512}} & {{0.494}} & {{0.532}} & {{0.466}} & {{0.462}} & {{0.353}} & {{0.469}} \\
\rowcolor{gray!20} MinCD  & \textcolor{blue}{\textbf{0.660}} & \textcolor{blue}{\textbf{0.642}} & {{0.536}} & \textcolor{blue}{\textbf{0.550}} & {{0.546}} & \textcolor{blue}{\textbf{0.471}} & \textcolor{blue}{\textbf{0.568}} \\
Hg-I2P$^\dag$  & {{0.526}} & {{0.521}} & {{0.537}} & {{0.486}} & {{0.491}} & {{0.408}} & {{0.494}} \\
Hg-I2P (Ours) & {{0.601}} & {{0.599}} & \textcolor{blue}{\textbf{0.585}} & {{0.544}} & \textcolor{blue}{\textbf{0.556}} & {{0.465}} & {{0.558}} \\
\hline
RR & O$\rightarrow$C & O$\rightarrow$F & O$\rightarrow$H & O$\rightarrow$P & O$\rightarrow$K &  O$\rightarrow$S & AVG \\
\hline
%P2-Net     & 0.566 & 0.661 & 0.232 & 0.577 & 0.532 & 0.234 & 0.510 \\
MATR  & 0.660 & 0.556 & 0.417 & 0.395 & 0.636 & 0.286 & 0.491 \\
Top-I2P  & {{0.702}} & {{0.759}} & {{0.423}} & {{0.605}} & {{0.645}} & {{0.293}} & {{0.571}} \\
MinCD   & {{0.769}} & {{0.726}} & {{0.250}} & {{0.681}} & \textcolor{blue}{\textbf{0.810}} & \textcolor{blue}{\textbf{0.643}} & {{0.647}} \\
Hg-I2P$^\dag$  & {{0.738}} & {{0.630}} & \textcolor{blue}{\textbf{0.750}} & \textcolor{blue}{\textbf{0.682}} & {{0.667}} & {{0.501}} & {{0.661}} \\
\rowcolor{gray!20} Hg-I2P (Ours) & \textcolor{blue}{\textbf{0.846}} & \textcolor{blue}{\textbf{0.781}} & {{0.583}} & {{0.638}} & {{0.768}} & {{0.523}} & \textcolor{blue}{\textbf{0.690}} \\
%\hline
\hline
IR & K$\rightarrow$C & K$\rightarrow$F  & K$\rightarrow$O & K$\rightarrow$H & K$\rightarrow$P & K$\rightarrow$S & AVG \\
\hline
P2-Net     & 0.516 & 0.512 & 0.504 & 0.506 & 0.555 & 0.358 & 0.491 \\
MATR  & 0.571 & 0.594 & 0.537 & 0.538 & 0.612 & 0.370 & 0.537 \\
Top-I2P & {{0.626}} & {{0.619}} & {{0.627}} & {{0.631}} & {{0.643}} & {{0.433}} & {{0.596}} \\
MinCD  & {{0.617}} & {{0.598}} & {{0.540}} & {{0.573}} & {{0.636}} & {{0.445}} & {{0.568}} \\
Hg-I2P$^\dag$  & {{0.645}} & {{0.677}} & {{0.651}} & {{0.602}} & {{0.686}} & {{0.448}} & {{0.618}} \\
\rowcolor{gray!20} Hg-I2P (Ours) & \textcolor{blue}{\textbf{0.728}} & \textcolor{blue}{\textbf{0.755}} & \textcolor{blue}{\textbf{0.705}} & \textcolor{blue}{\textbf{0.675}} & \textcolor{blue}{\textbf{0.753}} & \textcolor{blue}{\textbf{0.511}} & \textcolor{blue}{\textbf{0.688}} \\
\hline
RR & K$\rightarrow$C & K$\rightarrow$F  & K$\rightarrow$O & K$\rightarrow$H & K$\rightarrow$P & K$\rightarrow$S & AVG \\
\hline
%P2-Net    & 0.857 & 0.583 & 0.250 & 0.769 & 0.611 & 0.429 & 0.621 \\
MATR & 0.872 & 0.778 & 0.667 & 0.723 & 0.698 & 0.500 & 0.706 \\
Top-I2P &  {{0.936}} &  {{0.792}} &  {{0.722}} &  {{0.831}} &  {{0.860}} &  {{0.571}} &  {{0.785}} \\
MinCD   & {{0.846}} & \textcolor{blue}{\textbf{0.904}} & {{0.683}} & {{0.798}} & \textcolor{blue}{\textbf{0.872}} & {{0.786}} & {{0.814}} \\
Hg-I2P$^\dag$  & {{0.923}} & {{0.849}} & {{0.750}} & {{0.859}} & {{0.787}} & {{0.643}} & {{0.802}} \\
\rowcolor{gray!20} Hg-I2P (Ours) &  \textcolor{blue}{\textbf{0.969}} &  {{0.890}} &  \textcolor{blue}{\textbf{0.833}} &  \textcolor{blue}{\textbf{0.869}} &  {{0.702}} &  \textcolor{blue}{\textbf{0.857}} &  \textcolor{blue}{\textbf{0.853}} \\
\hline
%\bottomrule
\end{tabular}}
\label{tab:res_1}
\end{table}

\begin{table}[t]
%\scriptsize
\centering
\caption{Cross-dataset I2P registration performance across multiple datasets, including RGBD-V2~\cite{RGBD-dataset}, ScanNet~\cite{scan-net}, and a self-collected dataset. }
\resizebox{1.0\linewidth}{!}{%}
\begin{tabular}{c|ccccccc|c}
\toprule
%\hline
IR & RG-1 & RG-2 & RG-3 & RG-4 & RG-5 & RG-6 & RG-7 & AVG \\
\hline
%P2-Net    & 0.090 & 0.259 & 0.268 & 0.263 & 0.098 & 0.106 & 0.094 & 0.168 \\
MATR & 0.351 & 0.353 & 0.336 & 0.316 & 0.250 & 0.209 & 0.222 & 0.291 \\
Top-I2P   & {{0.397}} & {{0.413}} & {{0.389}} & {{0.398}} & {{0.252}} & {{0.247}} & {{0.253}} & {{0.335}} \\
MinCD   & {{0.427}} & {{0.415}} & {{0.405}} & {{0.412}} & {{0.310}} & \textcolor{blue}{\textbf{0.296}} & \textcolor{blue}{\textbf{0.329}} & {{0.371}} \\
\rowcolor{gray!20} Hg-I2P (Ours) & \textcolor{blue}{\textbf{0.503}} & \textcolor{blue}{\textbf{0.499}} & \textcolor{blue}{\textbf{0.444}} & \textcolor{blue}{\textbf{0.456}} & \textcolor{blue}{\textbf{0.325}} & {{0.285}} & {{0.314}} & \textcolor{blue}{\textbf{0.404}} \\
\hline
RR & RG-1 & RG-2 & RG-3 & RG-4 & RG-5 & RG-6 & RG-7 & AVG \\
\hline
%P2-Net    & 0.234 & 0.861 & 0.845 & 0.921 & 0.280 & 0.232 & 0.167 & 0.505 \\
MATR & 0.970 & {{0.880}} & 0.871 & 0.741 & 0.480 & 0.449 & 0.458 & 0.692 \\
Top-I2P   & {{0.973}} & 0.840 & \textcolor{blue}{\textbf{0.986}} & {{0.962}} & {{0.600}} & {{0.710}} & {{0.479}} & {{0.793}} \\
\rowcolor{gray!20} MinCD   & \textcolor{blue}{\textbf{0.974}} & \textcolor{blue}{\textbf{0.985}} & {{0.968}} & \textcolor{blue}{\textbf{0.963}} & {{0.707}} & {{0.725}} & {{0.711}} & \textcolor{blue}{\textbf{0.870}} \\
Hg-I2P (Ours) & {{0.966}} & 0.846 & {{0.972}} & {{0.926}} & \textcolor{blue}{\textbf{0.840}} & \textcolor{blue}{\textbf{0.754}} & \textcolor{blue}{\textbf{0.771}} & {{0.868}} \\
%\hline
\hline
IR & SC-1 & SC-2 & SC-3 & SC-4 & SC-5 & SC-6 & SC-7 & AVG \\
\hline
%P2-Net    & 0.359 & 0.413 & 0.286 & 0.190 & 0.370 & 0.312 & 0.195 & 0.303 \\
MATR & 0.495 & 0.550 & 0.424 & 0.337 & 0.507 & 0.434 & 0.414 & 0.451 \\
Top-I2P   & {{0.550}} & {{0.587}} & {{0.472}} & {{0.340}} & {{0.547}} & {{0.485}} & {{0.475}} & {{0.493}} \\
MinCD   & {{0.517}} & {{0.527}} & {{0.460}} & {{0.343}} & {{0.548}} & {{0.456}} & {{0.428}} & {{0.468}} \\
\rowcolor{gray!20} Hg-I2P (Ours) & \textcolor{blue}{\textbf{0.626}} & \textcolor{blue}{\textbf{0.691}} & \textcolor{blue}{\textbf{0.580}} & \textcolor{blue}{\textbf{0.453}} & \textcolor{blue}{\textbf{0.640}} & \textcolor{blue}{\textbf{0.569}} & \textcolor{blue}{\textbf{0.541}} & \textcolor{blue}{\textbf{0.586}} \\
\hline
RR & SC-1 & SC-2 & SC-3 & SC-4 & SC-5 & SC-6 & SC-7 & AVG \\
\hline
%P2-Net    & 0.778 & 0.750 & 0.846 & 0.273 & 0.893 & 0.929 & 0.510 & 0.711 \\
MATR & {{0.956}} & {{0.954}} & 0.974 & 0.433 & 0.923 & 0.909 & 0.750 & 0.842 \\
Top-I2P   & 0.889 & 0.902 & {{0.976}} & {{0.620}} & {{0.962}} & \textcolor{blue}{\textbf{0.981}} & \textcolor{blue}{\textbf{0.976}} & {{0.901}} \\
MinCD   & {{0.987}} & \textcolor{blue}{\textbf{0.979}} & {{0.905}} & \textcolor{blue}{\textbf{0.720}} & {{0.962}} & {{0.915}} & {{0.821}} & {{0.898}} \\
\rowcolor{gray!20} Hg-I2P (Ours) & \textcolor{blue}{\textbf{0.996}} & 0.944 & \textcolor{blue}{\textbf{0.986}} & {{0.705}} & \textcolor{blue}{\textbf{0.968}} & {{0.912}} & {{0.944}} & \textcolor{blue}{\textbf{0.922}} \\
%\hline
\hline
IR & SE-1 & SE-2 & SE-3 & SE-4 & SE-5 & SE-6 & SE-7 & AVG \\
\hline
%P2-Net    & 0.336 & 0.299 & 0.270 & 0.429 & 0.304 & 0.443 & 0.235 & 0.331 \\
MATR & 0.497 & 0.462 & 0.426 & \textcolor{blue}{\textbf{0.618}} & {{0.507}} & \textcolor{blue}{\textbf{0.619}} & 0.412 & 0.506 \\
Top-I2P   & {{0.506}} & {{0.499}} & {{0.476}} & 0.617 & 0.473 & 0.615 & {{0.459}} & {{0.521}} \\
MinCD   & {{0.485}} & {{0.470}} & {{0.437}} & {{0.581}} & {{0.522}} & {{0.604}} & {{0.514}} & {{0.516}} \\
\rowcolor{gray!20} Hg-I2P (Ours) & \textcolor{blue}{\textbf{0.702}} & \textcolor{blue}{\textbf{0.615}} & \textcolor{blue}{\textbf{0.649}} & 0.460 & \textcolor{blue}{\textbf{0.730}} & 0.366 & \textcolor{blue}{\textbf{0.589}} & \textcolor{blue}{\textbf{0.573}} \\
\hline
RR & SE-1 & SE-2 & SE-3 & SE-4 & SE-5 & SE-6 & SE-7 & AVG \\
\hline
%P2-Net    & 0.333 & 0.224 & 0.056 & 0.833 & 0.222 & 0.942 & 0.052 & 0.380 \\
MATR & 0.556 & 0.389 & 0.333 & {{0.976}} & 0.532 & \textcolor{blue}{\textbf{0.964}} & 0.278 & 0.575 \\
Top-I2P   & {{0.722}} & {{0.444}} & {{0.452}} & 0.944 & {{0.611}} & 0.952 & {{0.333}} & {{0.636}} \\
MinCD   & {{0.512}} & {{0.405}} & {{0.389}} & \textcolor{blue}{\textbf{0.984}} & {{0.611}} & {{0.952}} & {{0.389}} & {{0.606}} \\
\rowcolor{gray!20} Hg-I2P (Ours) & \textcolor{blue}{\textbf{0.889}} & \textcolor{blue}{\textbf{0.778}} & \textcolor{blue}{\textbf{0.772}} & 0.833 & \textcolor{blue}{\textbf{0.778}} & 0.871 & \textcolor{blue}{\textbf{0.556}} & \textcolor{blue}{\textbf{0.775}} \\
%\hline
\bottomrule
\end{tabular}}
\label{tab:res_2}
\end{table}

\section{Experiments and Discussions}
\label{sec.exp}

\subsection{Configurations} 
\label{sec.exp.A}

\begin{table}[t]
\centering
\caption{Comparisons of I2P registration methods on the ScanNet dataset~\cite{scan-net}. $\dag$ denotes evaluation with an RR threshold of 30 cm, while other methods use 5 cm.}
\resizebox{0.62\linewidth}{!}{%}
\begin{tabular}{c|c|c|c}
% \toprule
\hline
{Methods} & {Venue} & IR & {RR} \\
\hline
LCD$^\dag$      & AAAI 2020 & 0.307 & N/A   \\
Glue$^\dag$     & CVPR 2020 & 0.183 & 0.065 \\
P2-Net & ICCV 2021 & 0.303 & 0.711 \\
MATR   & ICCV 2023 & 0.451 & 0.842 \\
FreeReg$^\dag$  & ICLR 2024 & 0.568 & 0.780 \\
Top-I2P       & IJCAI 2025 & 0.493 & 0.901 \\
MinCD         & ICCV  2025 & 0.464 & 0.898 \\
\rowcolor{gray!20} Hg-I2P (Ours)        &          & \textcolor{blue}{\textbf{0.586}} & \textcolor{blue}{\textbf{0.922}}\\
\hline
%\bottomrule
\end{tabular}}
\label{tab:res_3}
\end{table}

% under cross-modal feature misalignment in unseen scenes
To evaluate the robustness and generalization capability of existing I2P registration methods, we conduct experiments primarily in a \textbf{cross-domain} configuration, which includes both cross-scene and cross-dataset evaluations. 

% datasets
\noindent\textbf{Datasets information}. Six datasets are employed, including five indoor datasets (i.e., 7-Scene~\cite{scene-7-dataset}, RGBD-V2~\cite{RGBD-dataset}, ScanNet~\cite{scan-net}, a self-collected dataset, and TUM~\cite{tum}), as well as one outdoor dataset, KITTI~\cite{kitti}. 

As I2P registration is mainly used for fine-grained visual localization within a pre-built point cloud map~\cite{ep2p-loc, p2-net}, each image maintains a high degree of overlap with the corresponding submap (i.e., a segmented portion of the map determined by a coarse camera pose). Thus, to simulate the realistic applications, we generate I2P samples with high overlaps. The 3D point clouds are down-sampled using a voxel resolution of 1.5 cm, with a maximum of $3.0\times 10^5$ points. RGB images are resized to $640\times 480$ resolution. Since KITTI images are originally $1242\times 375$, we crop and pad them to $640\times 480$ and adjust the camera intrinsic parameters accordingly. As LiDAR scans are typically sparser than depth camera point clouds, we generate LiDAR point clouds through the multi-scan registration. Further details of the self-collected dataset are provided in {the supplemental materials}. All point clouds in our experiments include RGB features.  

% compared methods and metrics
\noindent\textbf{Baseline methods and evaluation metrics}. Top-I2P~\cite{top-i2p} and MinCD~\cite{mincd} represent the current state-of-the-art methods for I2P registration in unseen scenes (more discussion is shown in the supplemental materials). Hg-I2P is compared against these methods under a cross-domain evaluation setting, where all models are trained in one scene and tested on the unseen scenes. Besides, Hg-I2P is also compared with more existing methods \cite{freereg, p2-net, graph-i2p-cvpr, corr-learning-i2p-cvpr, bridge} in the standard evaluation setting. During the evaluation, inlier ratio (IR) and registration recall (RR) are main metrics~\cite{2d3d-matr}. In addition, Average relative translational error (RTE) and average relative rotation error (RRE)~\cite{CoFiI2P,corr-tcsvt} are used for further analysis. 

% training scheme and hyper-parameters
\noindent\textbf{Implementation details}. Before registration, the pre-built map is segmented using SAM3D~\cite{sam3d_used}, while FastSAM~\cite{fastsam} is used for image segmentation during inference because of its computational efficiency. Using GT camera poses, we align 2D and 3D segments via Eq.~\ref{eq_3} to generate the GT heterogeneous edges $\mathbf{{E}}_{\text{I2P}}$. The backbone networks in Hg-I2P are adapted from from MATR~\cite{2d3d-matr}, and we adopt the same learning rate and optimizer as Top-I2P~\cite{top-i2p}. Training is conducted for 60 epochs to ensure accurate heterogeneous edge learning. The main hyperparameters in Hg-I2P (i.e., $\alpha$, $\delta_{\text{rej}}$, $\tau_{\text{rej}}$, and $\lambda_1$) are discussed in the {supplemental materials}.  

\begin{table}[t]
%\scriptsize
\centering
\caption{Performance comparison on the TUM dataset~\cite{tum} (RR threshold is 10 cm). Hg-I2P consistently achieves the highest IR and RR, and lowest RTE and RRE across all sequences.}
\resizebox{0.7\linewidth}{!}{%}
\begin{tabular}{c|c|c|c|c}
% \toprule
\hline
{Freiburg\_01} & IR & RR & {RTE/m} & {RRE/deg} \\
\hline
MATR & 0.319 & 0.595 & 0.037 & 1.238 \\
Top-I2P & 0.306 & 0.655 & 0.036 & 1.179 \\
\rowcolor{gray!20} Hg-I2P  (Ours)       & \textcolor{blue}{\textbf{0.380}} & \textcolor{blue}{\textbf{0.681}} & \textcolor{blue}{\textbf{0.035}} & \textcolor{blue}{\textbf{1.147}}\\
\hline
{Freiburg\_02} & IR & RR & {RTE/m} & {RRE/deg} \\
\hline
MATR & 0.223 & 0.223 & 0.052 & 1.469 \\
Top-I2P & 0.234 & 0.310 & 0.054 & 1.502 \\
\rowcolor{gray!20} Hg-I2P (Ours)        & \textcolor{blue}{\textbf{0.286}} & \textcolor{blue}{\textbf{0.394}} & \textcolor{blue}{\textbf{0.050}} & \textcolor{blue}{\textbf{1.455}}\\
\hline
{Freiburg\_03} & IR & RR & {RTE/m} & {RRE/deg} \\
\hline
MATR & 0.335 & 0.548 & 0.051 & 1.310 \\
Top-I2P & 0.314 & 0.603 & 0.050 & 1.294 \\
\rowcolor{gray!20} Hg-I2P (Ours)        & \textcolor{blue}{\textbf{0.393}} & \textcolor{blue}{\textbf{0.666}} & \textcolor{blue}{\textbf{0.047}} & \textcolor{blue}{\textbf{1.198}}\\
\hline
% \bottomrule
\end{tabular}}
\label{tab:res_6}
\end{table}

\subsection{Comparisons} 
\label{sec.exp.B}

We conduct cross-domain I2P registration evaluations, including both \textbf{cross-scene} and \textbf{cross-dataset} comparisons. 
% Additional visualizations and analyses are provided in {the supplemental materials}.

\vspace{+1mm}
\noindent\textbf{Cross-scene comparisons.} This experiment is conducted on the 7-Scenes dataset~\cite{scene-7-dataset}, which includes the scenes Chess (C), Fire (F), Heads (H), Office (O), Pumpkin (P), Kitchen (K), and Stairs (S). For simplicity, we use C$\rightarrow$F to indicate that the model is trained in scene C and evaluated on scene F. Table~\ref{tab:res_1} presents the comparison results with RR threshold as 2.5 cm. Even without the HC-pruning module, the proposed Hg-I2P outperforms MATR~\cite{2d3d-matr}, Top-I2P~\cite{top-i2p}, and MinCD~\cite{mincd} on both IR and RR metrics across most scenes. Hg-I2P (w/o HC-pruning) shows strong performance because accurate heterogeneous edge prediction enhances cross-modal feature interaction, thereby improving feature discriminability. When HC-pruning is applied, IR and RR increase further, as unreliable correspondences are effectively removed by proposed pruning criteria. These results collectively confirm the effectiveness of Hg-I2P. 

\begin{figure}[t]
	\centering
		\includegraphics[width=1.0\linewidth]{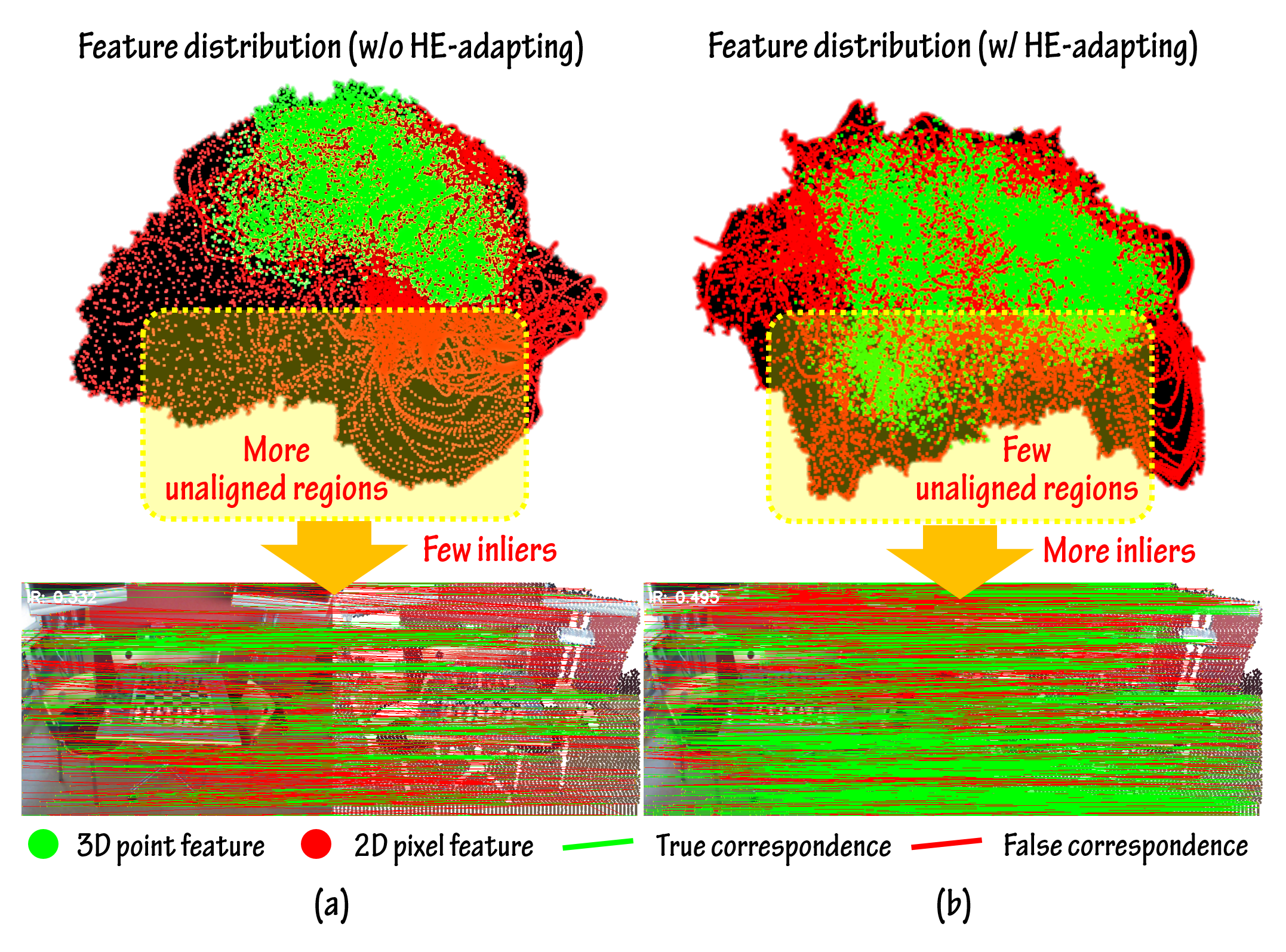}	
		\caption{Effect of the proposed HE-adapting module. (a) Baseline without HE-adapting. (b) With HE-adapting, cross-modal features are more accurately aligned, producing a higher number of inliers.}
	\label{fig:supp_vis_fea}
\end{figure}

\begin{table}[t]
%\scriptsize
\centering
\caption{Comparisons on the KITTI dataset~\cite{kitti}. $\dag$ indicates that LiDAR point clouds are constructed from multiple scans, a common setup in visual localization using pre-built maps.}
\resizebox{0.725\linewidth}{!}{%}
\begin{tabular}{c|c|c|c}
% \toprule
\hline
{Methods} & {Venue} & {RTE/m} & {RRE/deg} \\
\hline
%DeepI2P & {CVPR 2021} & 1.460 & 4.270 \\
CorrI2P & {TCSVT 2022} & 0.740 & 2.070 \\
VP2P-match & {NeurIPS 2023} & 0.750 & 3.290 \\
CoFiI2P & {RAL 2024} & 0.290 & 1.140 \\
CMR-Agent & {IROS 2024} & 0.195 & 0.589 \\
GraphI2P & {CVPR 2025}  & 0.320 & 1.650 \\
ImCorr   & {CVPR 2025}  & 0.200 & 1.240 \\
Hg-I2P (Ours)  &              & 0.172 & 0.502 \\
\rowcolor{gray!20} Hg-I2P$^\dag$  (Ours)        &          & \textcolor{blue}{\textbf{0.160}} & \textcolor{blue}{\textbf{0.485}}\\
\hline
% \bottomrule
\end{tabular}}
\label{tab:res_5}
\end{table}

\begin{table}[t]
%\scriptsize
\centering
\caption{Comparisons on the RGBD-V2 dataset~\cite{RGBD-dataset}. $\dag$ denotes methods requiring depth input from RGB-D camera.}
\resizebox{0.6\linewidth}{!}{%}
\begin{tabular}{c|c|c|c}
% \toprule
\hline
{Methods} & {Venue} & IR & {RR} \\
\hline
%FCGF & ICCV 2019 & 0.081 & 0.304 \\
%Predator & CVPR 2020 & 0.157 & 0.302 \\
P2-Net & ICCV 2021 & 0.122 & 0.384 \\
MATR   & ICCV 2023 & 0.324 & 0.564 \\
FreeReg  & ICLR 2024 & 0.309 & 0.573 \\
DiffReg$^\dag$  & ECCV 2024 & 0.378 & \textcolor{blue}{\textbf{0.874}} \\
Bridge   & AAAI 2025 & 0.351 & 0.634 \\
DiffI2P  & ICCV 2025 & 0.369 & 0.605 \\
FlowI2P  & IJCV 2025 & 0.401 & 0.684 \\
% MinCD    & ICCV 2025 & \\
\rowcolor{gray!20} Hg-I2P (Ours)        &          & \textcolor{blue}{\textbf{0.428}} & {{0.690}}\\
\hline
% \bottomrule
\end{tabular}}
\label{tab:res_4}
\end{table}

\vspace{+1mm}
\noindent\textbf{Cross-dataset comparisons without fine-tuning.} This evaluation is performed on multiple datasets, including RGBD-V2~\cite{RGBD-dataset}, ScanNet~\cite{scan-net},  self-collected, and TUM~\cite{tum} datasets. Methods are trained on scene K of the 7-Scenes dataset~\cite{scene-7-dataset} and directly tested on these unseen datasets without fine-tuning. Notations RG-X, SC-X, and SE-X denote the X-th scene from RGBD-V2~\cite{RGBD-dataset}, ScanNet~\cite{scan-net}, and the self-collected dataset, respectively. Table~\ref{tab:res_2} summarizes the results (RR threshold is 10 cm for RGBD-V2 and 5 cm for ScanNet and self-collected datasets). Hg-I2P achieves stable and high RR scores across all datasets. Additional comparisons with LCD~\cite{lcd}, Glue~\cite{superglue}, and FreeReg~\cite{freereg} on ScanNet (Table~\ref{tab:res_3}) further demonstrate Hg-I2P's robustness. Results on the TUM dataset (Table~\ref{tab:res_6}) show that Hg-I2P achieves RTE below 5 cm, outperforming prior methods~\cite{2d3d-matr, top-i2p}. Overall, Hg-I2P consistently delivers accurate registration in unseen domains without any fine-tuning. 

\vspace{+1mm}
\noindent\textbf{Cross-dataset comparisons with fine-tuning.} This experiment is conducted on the KITTI dataset~\cite{kitti}. Since LiDAR point clouds exhibit different density and range characteristics from depth-camera point clouds, Hg-I2P is fine-tuned on KITTI. As shown in Table~\ref{tab:res_5}, Hg-I2P outperforms existing methods, including CorrI2P~\cite{corr-tcsvt}, VP2P-match~\cite{diff_reg_match}, CoFiI2P~\cite{CoFiI2P}, CMR-Agent~\cite{cmr-agent}, GraphI2P~\cite{graph-i2p-cvpr}, and ImCorr~\cite{corr-learning-i2p-cvpr}, on both RTE and RRE metrics. These results confirm that the proposed MP-mining, HE-adapting, and HC-pruning modules are highly compatible with outdoor scenes and diverse sensor configurations. 

\vspace{+1mm}
\noindent\textbf{Benchmark comparisons.} We further evaluate Hg-I2P on the RGBD-V2 dataset~\cite{RGBD-dataset} following the standard training and testing split~\cite{2d3d-matr} (RR threshold is 10 cm). As shown in Table~\ref{tab:res_4}, Hg-I2P achieves superior IR and RR performance compared with state-of-the-art works~\cite{freereg, diff-reg, bridge, diffi2p, flowi2p}. This demonstrates that Hg-I2P effectively reduces the cross-modality gap between 2D images and 3D point clouds more efficiently than existing methods.  

\begin{figure}[t]
	\centering
		\includegraphics[width=1.0\linewidth]{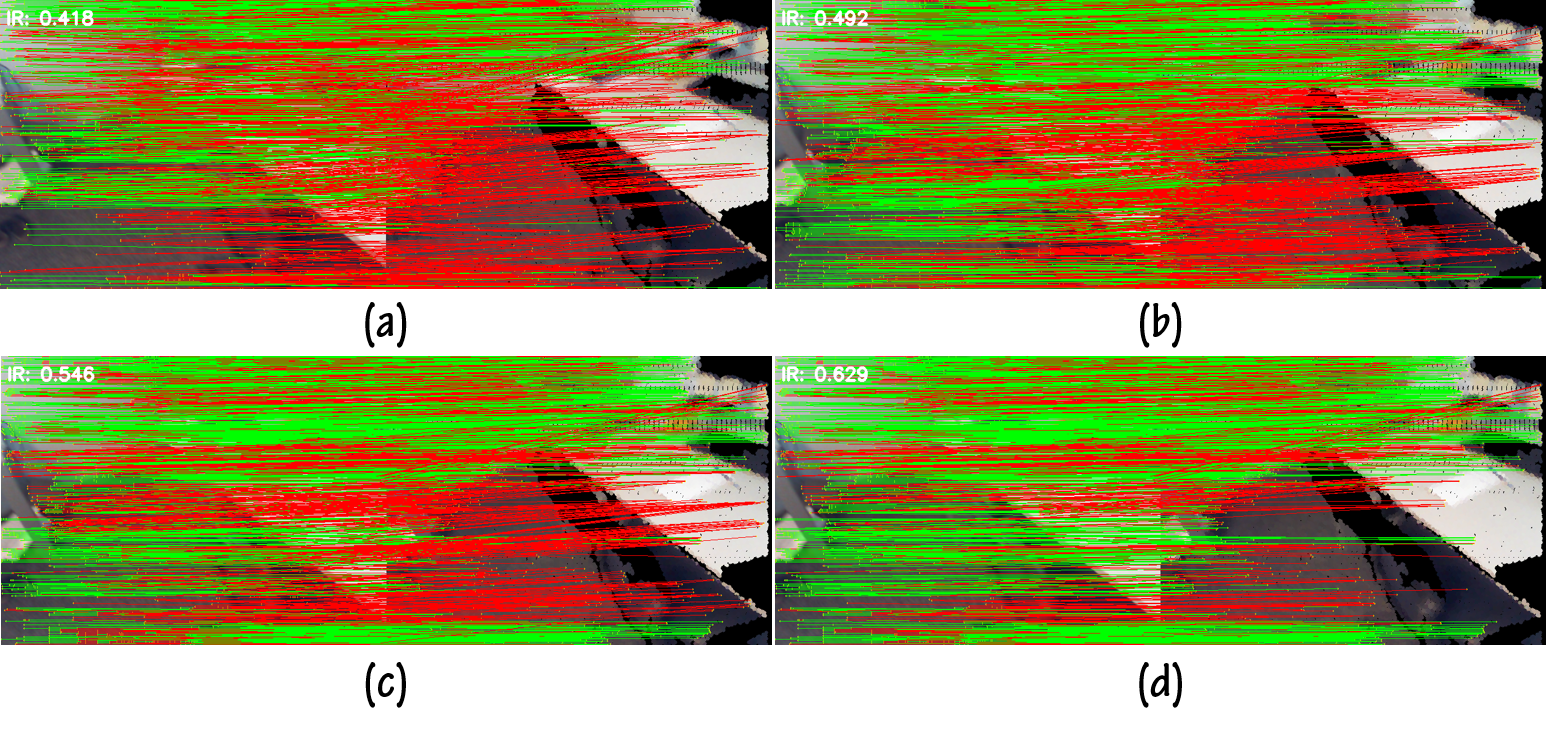}	
		\caption{Qualitative comparison of I2P registration results. (a) MATR~\cite{2d3d-matr}. (b) Top-I2P~\cite{top-i2p}. (c) Hg-I2P without HC-pruning. (d) Full Hg-I2P which produces fewer outliers than prior methods.}
	\label{fig:vis_ab}
\end{figure}

\begin{table}[t]
\centering
\caption{Ablation study of the MP-mining module. $\mathbf{E}_{\text{I2P}}^i$ denotes the inclusion of the $i$-th feature heterogeneous edge prediction.}
\resizebox{1.0\linewidth}{!}{%}
\begin{tabular}{ccc|c|c|c}
% \toprule
\hline
$\mathbf{E}_{\text{I2P}}^1$ & $\mathbf{E}_{\text{I2P}}^2$ & $\mathbf{E}_{\text{I2P}}^3$ & Precision  & IR & {RR} \\
\hline
$\checkmark$  &  &  & 0.826 & 0.478 & 0.610    \\
$\checkmark$  & $\checkmark$ &  & 0.874 & 0.490 & 0.634    \\
\rowcolor{gray!20} $\checkmark$  & $\checkmark$ & $\checkmark$ & 0.893 (\textcolor{blue}{\textbf{+6.7\%}}) & 0.494 (\textcolor{blue}{\textbf{+1.6\%}}) & 0.661 (\textcolor{blue}{\textbf{+5.1\%}})    \\
\hline
%\bottomrule
\end{tabular}}
\label{tab:ab_1}
\end{table}

\begin{table}[t]
\centering
\caption{Ablation study of the HE-adapting module. Parameter $\beta$ controls the strength of cross-modal feature interaction through heterogeneous edges.}
\resizebox{1.0\linewidth}{!}{%}
\begin{tabular}{c|cc>{\columncolor{gray!20}}cccc}
% \toprule
\hline
Parameter $\beta$ in Eq. (8) & 0.0 & 0.05 & 0.10 & 0.15 & 0.20 & 0.25 \\
\hline
IR  & 0.462 & 0.478 & \textcolor{blue}{\textbf{0.494}} & 0.482 & 0.454 & 0.437   \\
RR  & 0.620 & 0.648 & \textcolor{blue}{\textbf{0.661}} & 0.612 & 0.603 & 0.582   \\
\hline
%\bottomrule
\end{tabular}}
\label{tab:ab_2}
\end{table}

\subsection{Ablation studies} 
\label{sec.exp.C}

We analyze the effectiveness of each proposed module in Hg-I2P through a series of ablation studies on the 7-Scenes dataset~\cite{scene-7-dataset}. The Office scene is used for training, while the remaining scenes are used for evaluation. Additional results on hyperparameter sensitivity and runtime are provided in {the supplemental materials}. 

\vspace{+1mm}
\noindent\textbf{Effectiveness of MP-mining.} We first verify the MP-mining module in heterogeneous edge prediction. Table~\ref{tab:ab_1} shows that multi-path relational features improves edge prediction precision, which in turn leads to higher IR and RR scores. This confirms that MP-mining effectively captures richer inter-region dependencies for cross-modal learning. 

\vspace{+1mm}
\noindent\textbf{Effectiveness of HE-adapting.} We next evaluate the HE-adapting module for cross-modal feature adaptation. Results in Table~\ref{tab:ab_2} show that when $\beta=0.1$, both IR and RR reach their peak values, indicating optimal interaction intensity between heterogeneous features. However, when $\beta$ increases further, over-interaction reduces feature discriminability, highlighting the importance of balanced feature exchange. We also visualize the cross-modal features in 3D space with principal component analysis (PCA), shown in Fig.~\ref{fig:supp_vis_fea}. This PCA visualization confirms that HE-adapting aligns 2D and 3D features more effectively, producing more accurate correspondences. 

\begin{table}[t]
\centering
\caption{Ablation study of the HC-pruning module. ``Criterion X'' indicates the use of the $X$-th HC-pruning criterion.}
\resizebox{0.85\linewidth}{!}{%}
\begin{tabular}{cc|c|c}
% \toprule
\hline
Criterion I & Criterion II & IR & {RR} \\
\hline
              &  & 0.494 & 0.661    \\
$\checkmark$  &  & 0.523 & 0.682    \\
\rowcolor{gray!20} $\checkmark$  & $\checkmark$ & 0.558 (\textcolor{blue}{\textbf{+6.4\%}}) & 0.690 (\textcolor{blue}{\textbf{+2.9\%}})   \\
\hline
%\bottomrule
\end{tabular}}
\label{tab:ab_3}
\end{table}

\vspace{+1mm}
\noindent\textbf{Effectiveness of HC-pruning.} We finally assess the HC-pruning module for correspondence refinement. Table~\ref{tab:ab_3} shows that applying the reprojection distance-based criterion brings limited improvement on IR and RR, since $\mathbf{\tilde{T}}$ is still inaccurate. When the relative-coordinate consistency based criterion is also included, false correspondences are more effectively removed, leading to significant gains in IR and RR. As visualized in Fig.~\ref{fig:vis_ab}, HC-pruning reduces false-positive matches and enhances registration robustness. 

\section{Conclusions}
\label{sec.conc}

In this paper, we proposed Hg-I2P, a heterogeneous graph-embedded image-to-point-cloud registration framework designed for robust and generalizable cross-modal alignment. The core idea is to introduce a heterogeneous graph that connects segmented 2D and 3D regions, enabling cross-modal feature interaction through heterogeneous edges and outlier suppression through graph-based projection consistency.
Extensive experiments across six datasets, including both indoor and outdoor scenarios, demonstrate that Hg-I2P significantly outperforms existing methods in terms of generalization and accuracy. %Limitations and future work of Hg-I2P are discussed in {the supplemental materials}.

% Hg-I2P thus provides a unified and effective solution for reliable 2D-3D registration across diverse real-world environments. 

\section*{Acknowledgments}

This work is partially supported by the National Key R\&D Program of China (Grant ID: 2024YFC3015303), National Natural Science Foundation of China (Grand ID: 62502171), and China Postdoctoral Science Foundation (Grand ID: 2024M761014 and GZC20252285).
\section*{Appendix A. More Details of the Self-Collected Dataset}

\begin{figure}[h]
	\centering
		\includegraphics[width=1.0\linewidth]{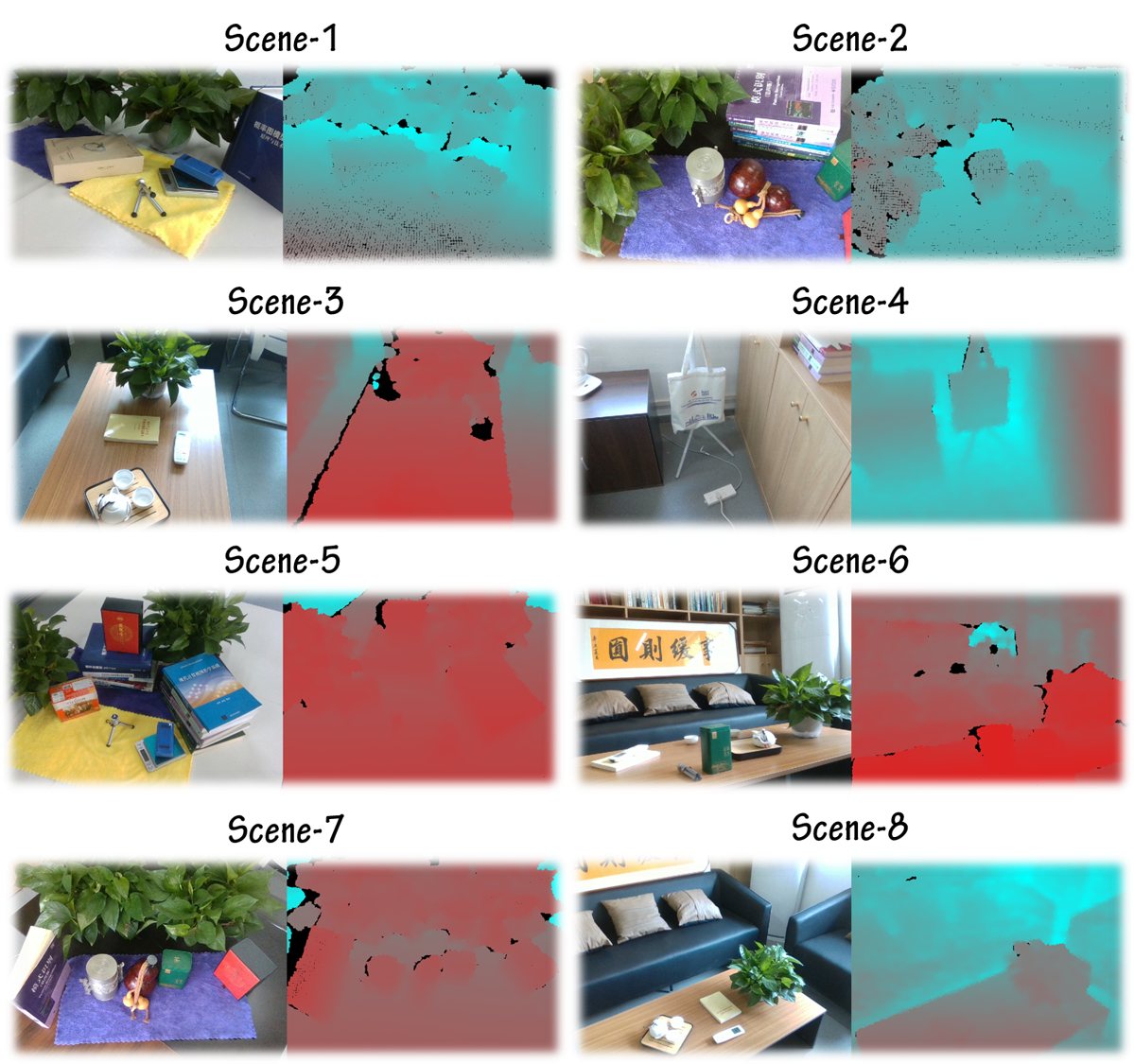}	
		\caption{Examples from the self-collected indoor I2P dataset.  Representative RGB and depth scenes captured using an Intel RealSense D415 in indoor environments such as tables, living rooms, and shelf-like spaces. The dataset includes realistic depth noise and fine-grained geometric variation and is used exclusively for zero-shot evaluation of I2P registration.}
	\label{fig:supp_dataset}
\end{figure}

To evaluate the generalization ability of I2P registration, we collected an I2P dataset on a campus building. Given that I2P registration is mainly used in AR/VR applications, this dataset was collected in representative indoor scenes, including tables, living rooms, paintings, books, and toys. RGB-D data was captured using an Intel RealSense depth camera \texttt{D415}, and images were resized to $640\times480$ resolution. The point clouds contain inherent noise caused by depth measurement errors, which is common in the real-world I2P dataset. Visualizations of this dataset are shown in Fig. \ref{fig:supp_dataset}. Ground truth (GT) camera poses are estimated using 3D Iterative Closest Point (ICP). For training, we also use RGB-D frames as samples, where the GT camera poses are set to the identity matrix. The depth range in this dataset varies from $1.0$ m to $9.0$ m. Because the dataset contains only 0.1K samples, it is used exclusively for zero-shot I2P registration.

\section*{Appendix B. More Implementation Details of Hg-I2P}

We provide additional technical details and motivations behind the implementation of Hg-I2P.

\vspace{+2mm}
\noindent\textbf{Segmentation}. To construct the heterogeneous graph, the first step is segmenting RGB images and 3D point clouds. A naive strategy is to use pre-trained semantic segmentation models. However, real-world indoor scenes often contain unseen or long-tail objects that such models fail to segment reliably. Therefore, we adopt the Segment Anything Model (SAM) \cite{sam_2d} to obtain fine-grained segmentation on RGB images. Because standard SAM requires nearly 1 second per image, we instead employ the pre-trained FastSAM \cite{fastsam} for faster inference.

\begin{figure*}[h]
	\centering
		\includegraphics[width=1.0\linewidth]{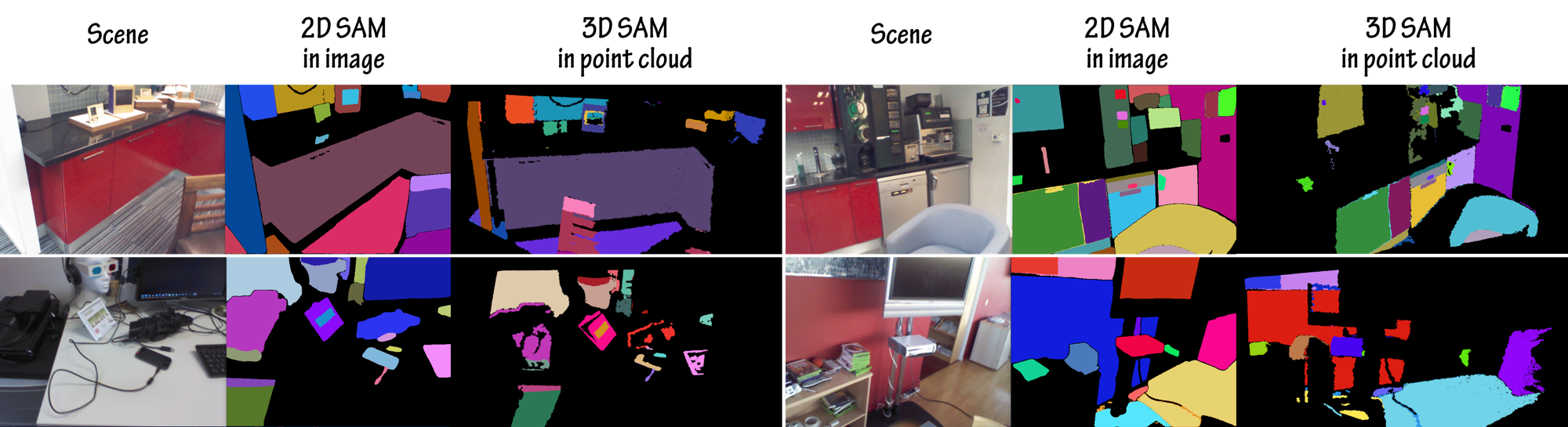}	
		\caption{2D and 3D SAM-based segmentation results. The resulting 2D and 3D region sets exhibit strong shape and boundary consistency, enabling reliable heterogeneous graph construction for cross-modal representation learning.}
	\label{fig:supp_segs}
\end{figure*}

To segment 3D point clouds, we follow Yang et al. \cite{sam3d_used}, which generates a segmented point-cloud map by fusing a sequence of SAM-segmented RGB-D frames. In I2P registration scenarios, point clouds are typically pre-built from RGB-D videos, making this fusion-based approach \cite{sam3d_used} especially suitable for 3D segmentation. Since both images and point clouds are segmented by SAM, 2D and 3D regions maintain strong shape consistency, which simplifies the prediction of heterogeneous edges. More visualizations of 2D and 3D segments are shown in Fig. \ref{fig:supp_segs}.  

\vspace{+2mm}
\noindent\textbf{Heterogeneous graph initialization}. We now describe the graph initialization process. Given an image $\mathcal{I}\in\mathbb{R}^{H\times W\times 3}$ and a point cloud $\mathcal{P}\in\mathbb{R}^{N\times 3}$, we extract a 2D feature map $\mathbf{F}_I\in\mathbb{R}^{H\times W\times C}$ using a ResNet-based U-Net following  \cite{2d3d-matr}. The $i$-th vertex feature $\mathbf{v}^I_i$ is computed as:
\begin{equation}
\mathbf{v}^I_i = \texttt{Torch.mean}(\mathbf{F}_I[\texttt{Mask}^I_i])
\end{equation}
\noindent Similarly, we extract 3D point features $\mathbf{F}_P\in\mathbb{R}^{N\times C}$ from $\mathcal{P}$ using a KPConv-based U-Net following \cite{2d3d-matr}, and compute the $j$-th vertex feature $\mathbf{v}^P_j$ as:
\begin{equation}
\mathbf{v}^P_j = \texttt{Torch.mean}(\mathbf{F}_P[\texttt{Mask}^P_j])
\end{equation}
We employ the same ResNet and KPConv backbones as in \cite{2d3d-matr} because of their efficiency and robustness, achieving a runtime of about 3 - 4 ms on an NVIDIA GTX 3080 GPU.  

\vspace{+2mm}
\noindent\textbf{Heterogeneous edge prediction}. As discussed in Sec. 5.1, I2P registration is primarily used for visual localization in the pre-built point cloud map, where a pose prior is available \cite{cmr-next}. This allows segmenting the relevant sub-map for I2P registration, and implies that the GT transform matrix $\mathbf{T}_{\text{gt}}$ is close to the identity matrix $\mathbf{I}$. In MP-mining, to enforce edge sparsity, we apply a mask $\mathbf{M}_{\text{I2P}} \in \mathbb{R}^{M\times N}$ to $\mathbf{\hat{E}}_{\text{I2P}}$ to remove incorrect edges based on a reprojection threshold $\tau_{2d}$. $\mathbf{\hat{E}}_{\text{I2P}}$ is computed as:
\begin{equation}
\mathbf{\hat{E}}_{\text{I2P}} \leftarrow \mathbf{\hat{E}}_{\text{I2P}} \odot \mathbf{M}_{\text{I2P}}, (\mathbf{M}_{\text{I2P}})_{ij} = \mathbf{1}(\Vert \bar{\mathbf{p}}^I_i - \bar{\mathbf{p}}^P_j \Vert_2 \leq \tau_{2d})
\end{equation}
\noindent where $\mathbf{1}(\cdot)$ is an indicator function. $\odot$ denotes an element-wise product. $\bar{\mathbf{p}}^I_i$ is the average pixel coordinate of the $i$-th image segment. $\bar{\mathbf{p}}^P_j$ is the projection of the center point of the $j$-th segment of the point cloud using the identity projection. $\tau_{2d}$ is empirically set to $16$. 

\vspace{+2mm}
\noindent\textbf{Criteria in HC-Pruning}. We next provide more details on HC-Pruning. To estimate the initial pose $\mathbf{\tilde{T}}$, we use a PnP-based cost over the centers of matched 2D-3D regions, i.e.,  $L_{\mathrm{pnp}}(\mathbf{\tilde{T}}) = \sum_{i, k\in \mathcal{E}_{\text{I2P}}(i|I)} \Vert \bar{\mathbf{p}}^I_i - \pi(\bar{\mathbf{P}}^P_k|\mathbf{\tilde{T}}) \Vert_2$, where $\bar{\mathbf{p}}^I_i$ and $\bar{\mathbf{P}}^P_k$ denote the respective centers of $\mathcal{I}_i$ and $\mathcal{P}_k$.
We then prune based on graph-aware criteria. Because the heterogeneous edge relations are effective at filtering outliers, our \textbf{first pruning criterion} identifies an inlier if at least one of the three geometric consistency conditions holds, i.e., \textit{$\langle \mathbf{p}_i^c, \mathbf{P}_i^c \rangle$ is an inlier if at least one of the conditions is satisfied: (i) $\mathbf{p}_i^c \in \mathcal{I}_i$, $\mathbf{P}_i^c \in \mathcal{P}_k, \forall k\in \mathcal{E}_{\text{I2P}}(i|I)$; (ii) $\mathbf{P}_i^c \in \mathcal{P}_i$, $\mathbf{p}_i^c \in \mathcal{I}_k, \forall k\in \mathcal{E}_{\text{I2P}}(i|P)$; (iii) $\Vert \mathbf{p}^c_i - \pi(\mathbf{P}^c_i|\mathbf{\tilde{T}}) \Vert_2 \leq \delta_{\text{rej}}$.} 

Afterward, we construct vectors $\mathbf{s}_i\in\mathbb{R}^M$ and $\mathbf{t}_i\in\mathbb{R}^N$ to describe each correspondence's relative position to graph vertices:
\begin{equation}
\begin{aligned}
\mathbf{s}_i&=(\mathbf{p}^c_i - \bar{\mathbf{p}}^I_1,...,\mathbf{p}^c_i - \bar{\mathbf{p}}^I_M)^T, 
\\
\mathbf{t}_i&=(\pi(\mathbf{P}^c_i|\mathbf{\tilde{T}})-\bar{\mathbf{q}}^I_1,...,\pi(\mathbf{P}^c_i|\mathbf{\tilde{T}})-\bar{\mathbf{q}}^I_M)^T
\end{aligned}
\end{equation}
\noindent where $(\bar{\mathbf{q}}^I_1,..., \bar{\mathbf{q}}^I_M)^T = \mathbf{\hat{E}}_{\text{I2P}} \cdot (\pi(\bar{\mathbf{P}}^P_1|\mathbf{\tilde{T}}),...,\pi(\bar{\mathbf{P}}^P_N|\mathbf{\tilde{T}}))^T$. In the ideal case (i.e., $\mathbf{\tilde{T}}$ and $\mathbf{\hat{E}}_{\text{I2P}}$ are correct), $\bar{\mathbf{q}}^I_i \approx \bar{\mathbf{p}}^I_i$ is satisfied. For an inlier, the cosine distance of $\mathbf{s}_i$ and $\mathbf{t}_i$ is close to $1$. Thus, the \textbf{second pruning criterion} identifies an inlier if this cosine distance meets a threshold, i.e., \textit{$\langle \mathbf{p}_i^c, \mathbf{P}_i^c \rangle$ is an inlier if the cosine distance of $\mathbf{s}_i$ and $\mathbf{t}_i$ is greater than $\tau_{\text{rej}}$.}
In practice, due to errors in $\mathbf{\tilde{T}}$ and $\mathbf{\hat{E}}_{\text{I2P}}$, we adopt a relaxed rule and treat a correspondence as an inlier if either criterion is satisfied.

\vspace{+2mm}
\noindent\textbf{Implementation details}. The model is trained using the Adam optimizer for 30 epochs with a batch size of 1, a learning rate of $10^{-4}$, and a weight decay of $10^{-6}$. All training and testing are conducted on a single NVIDIA GTX 3080 GPU. Detailed parameters will be provided in the open-source release. Training from scratch requires approximately 6 hours, while fine-tuning requires 2-3 hours. 

\section*{Appendix C. More Discussions of Method Comparisons}

We provide additional analysis of method comparisons.

\vspace{+2mm}
\noindent\textbf{Selection of compared methods}. Top-I2P \cite{top-i2p} (IJCAI'25) and MinCD \cite{mincd} (ICCV'25) are representative recent works focusing on I2P registration in open-domain scenarios. They outperform previous methods such as MATR \cite{2d3d-matr}, P2-Net \cite{p2-net}, FreeReg \cite{freereg}, and Bridge \cite{bridge}. Therefore, we focus our comparison on Top-I2P, MinCD, and the proposed Hg-I2P.

\begin{table}[t]
%\scriptsize
\centering
\caption{Mean RTE and RRE of MATR, Top-I2P, MinCD, and Hg-I2P in the cross-scene setting on 7-Scenes. The variant Hg-I2P$^\dag$ excludes HC-pruning. The results highlight the impact of heterogeneous graph modeling and pruning on registration accuracy.}
\resizebox{0.715\linewidth}{!}{%}
\begin{tabular}{c|c|c|c}
% \toprule
\hline
{Methods} & {Venue} & {RTE/m} & {RRE/deg} \\
\hline
MATR    & {ICCV  2023} & 0.029 & 0.955 \\
Top-I2P & {IJCAI 2025} & 0.030 & 0.963 \\
MinCD   & {ICCV  2025} & 0.027 & 0.937 \\
Hg-I2P  &              & 0.028 & 0.954 \\
\rowcolor{gray!20} Hg-I2P$^\dag$         &          & \textcolor{blue}{\textbf{0.026}} & \textcolor{blue}{\textbf{0.903}}\\
\hline
% \bottomrule
\end{tabular}}
\label{tab:supp_exp_1}
\end{table}

\begin{figure*}[h]
	\centering
		\includegraphics[width=1\linewidth]{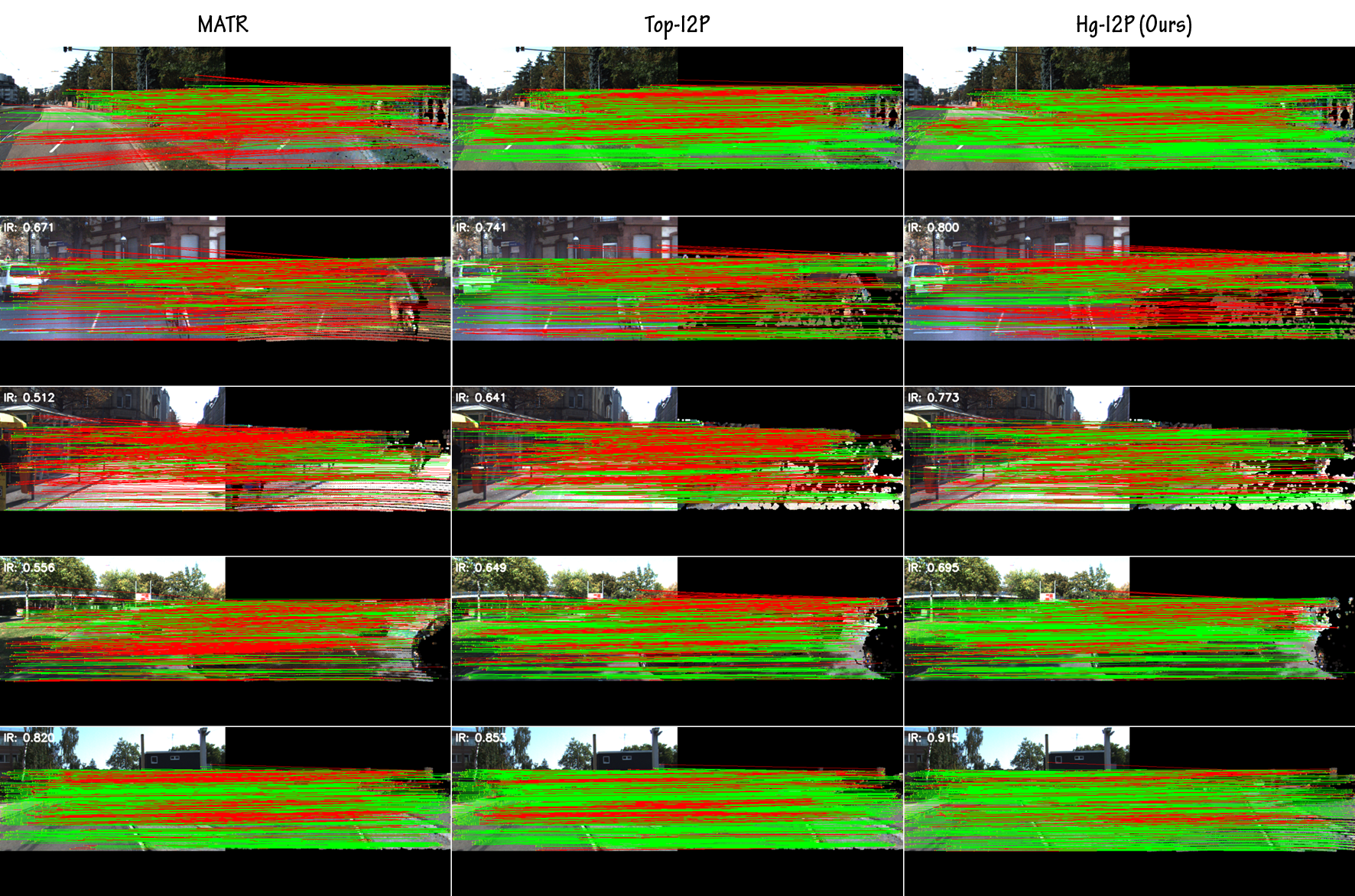}	
		\caption{Qualitative I2P registration results of MATR \cite{2d3d-matr} (left), Top-I2P \cite{top-i2p} (middle), and the proposed Hg-I2P (right) on the KITTI dataset \cite{kitti}. Hg-I2P produces more accurate and stable 2D-3D correspondences across diverse traffic scenes, benefiting from heterogeneous graph modeling and HC-pruning. Green and red lines denote correct and incorrect correspondences, respectively.}
	\label{fig:supp_kitti}
\end{figure*}

\begin{figure*}[h]
	\centering
		\includegraphics[width=1.0\linewidth]{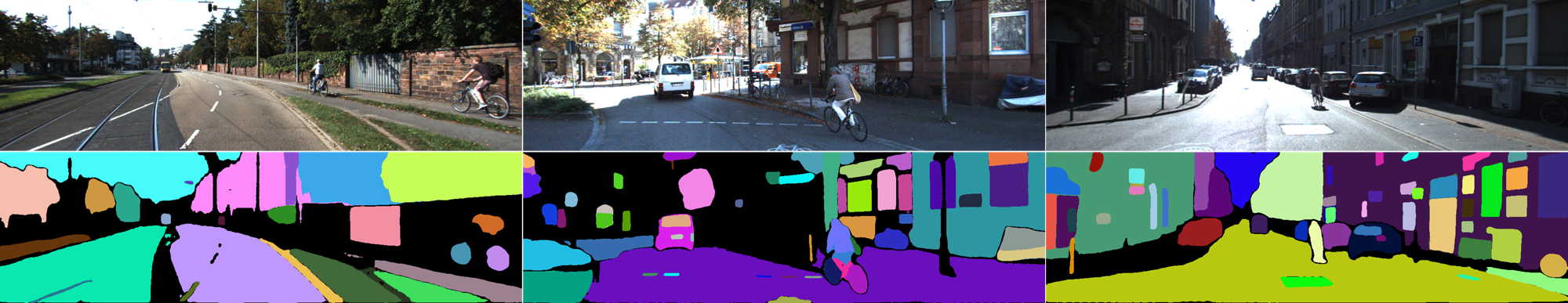}	
		\caption{Examples of SAM segmentation results on outdoor images from KITTI \cite{kitti}. Despite the presence of objects with irregular structures and large depth variability, SAM reliably identifies fine-grained regions, contributing to the robustness of Hg-I2P in outdoor environments after overlap extraction.}
	\label{fig:supp_kitti_sam}
\end{figure*}

\begin{figure*}[h]
	\centering
		\includegraphics[width=1.0\linewidth]{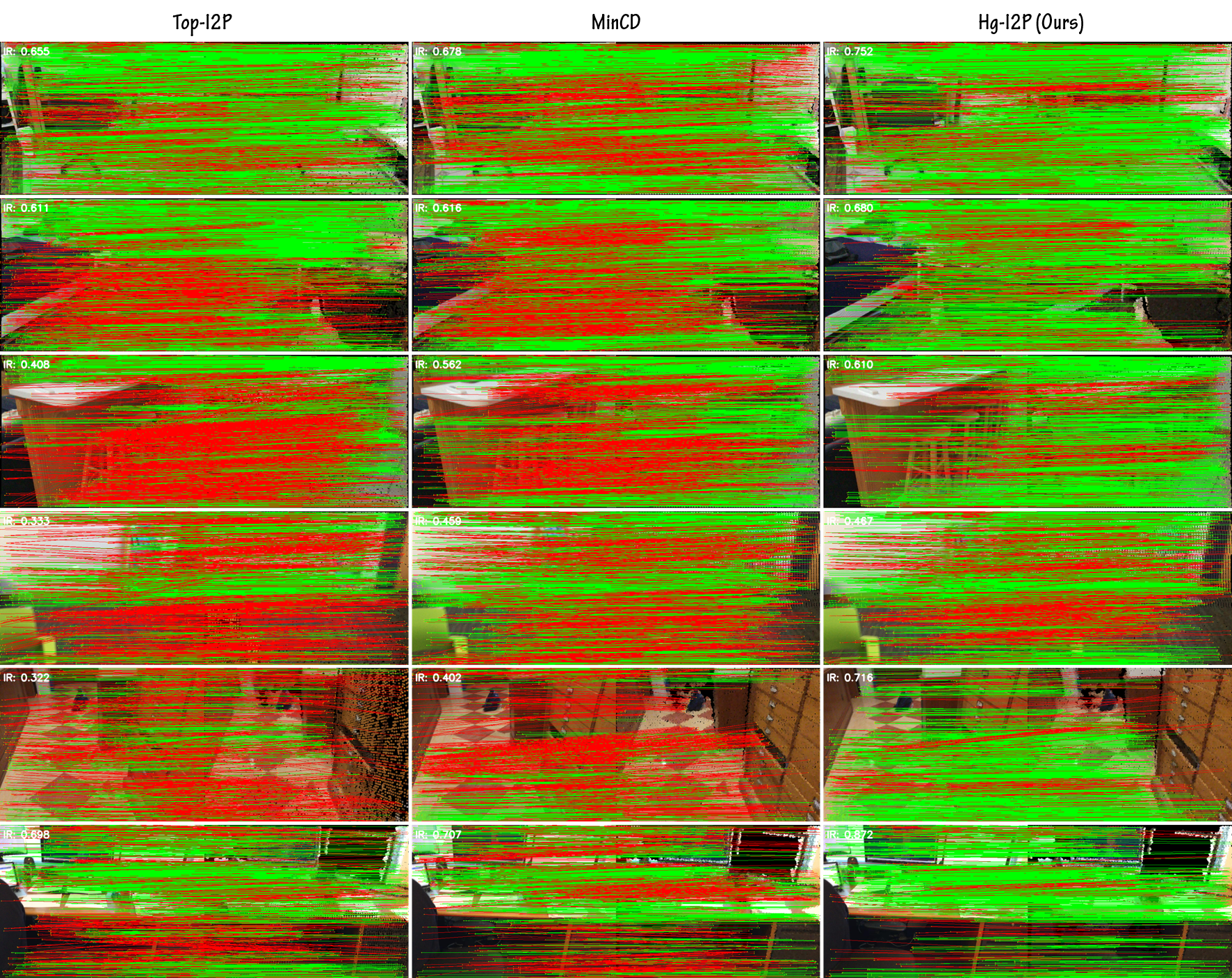}	
		\caption{Qualitative comparison of Top-I2P \cite{top-i2p} (left), MinCD \cite{mincd}, and Hg-I2P (right, ours) on ScanNet \cite{scan-net}. Hg-I2P consistently yields denser correct matches and significantly fewer outliers because of its heterogeneous graph design and multi-path relation mining. Green and red lines denote the correct and incorrect correspondences, respectively.}
	\label{fig:supp_qc_2}
\end{figure*}

\begin{figure*}[h]
	\centering
		\includegraphics[width=1.0\linewidth]{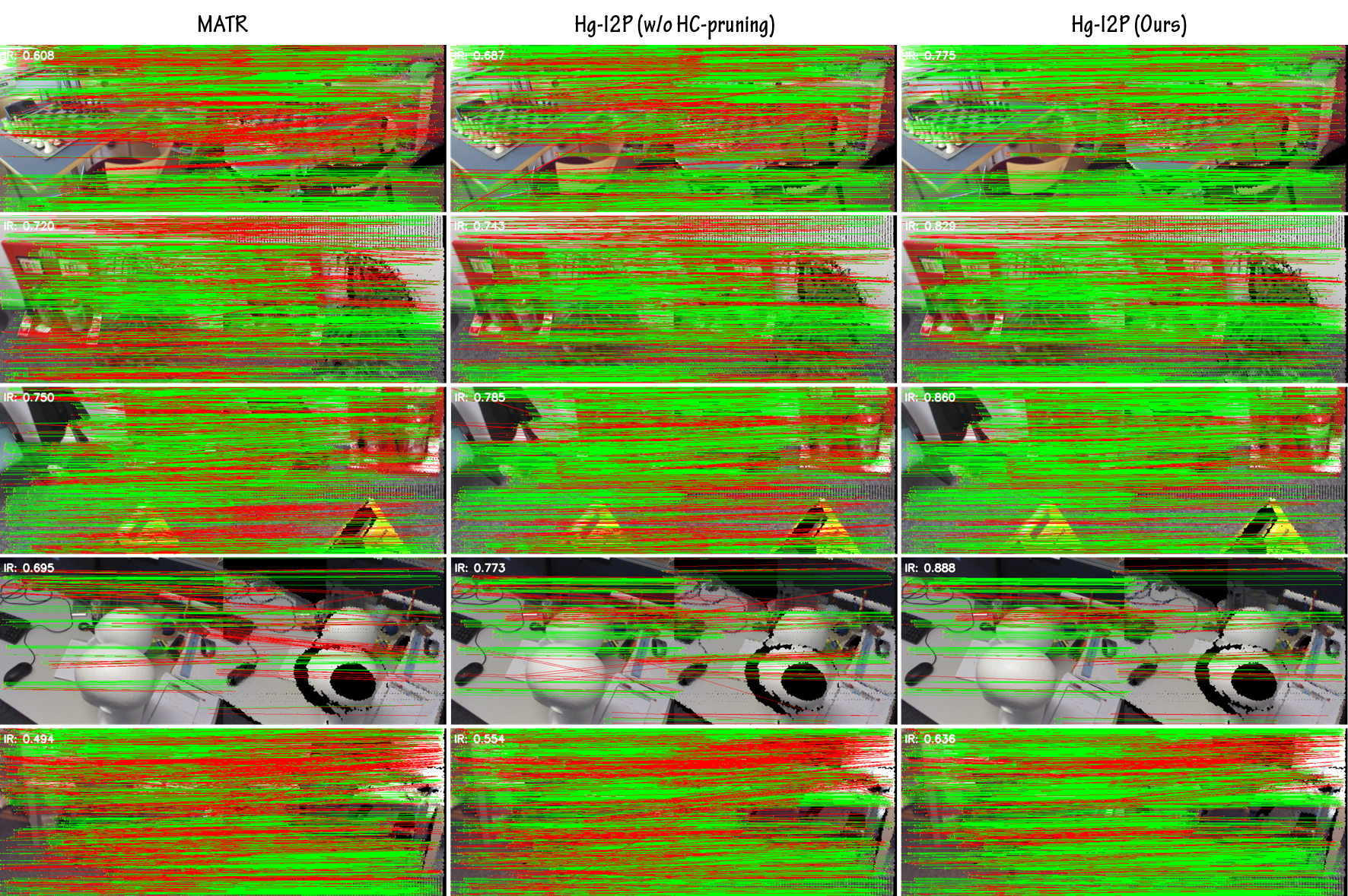}	
		\caption{Qualitative comparison among MATR \cite{2d3d-matr} (left), Hg-I2P without HC-pruning (middle), and Hg-I2P (right) on the 7-Scenes dataset \cite{kitti}. HC-pruning noticeably reduces incorrect correspondences and enhances the inlier ratio by enforcing graph-aware geometric consistency. Green and red lines denote correct and incorrect correspondences, respectively.}
	\label{fig:supp_qc_1}
\end{figure*}

\vspace{+2mm}
\noindent\textbf{In-depth analysis in indoor scenes}. Top-I2P \cite{top-i2p} and Hg-I2P share a common trait: \textit{both leverage SAM-segmented 2D and 3D regions for I2P registration}. However, Hg-I2P yields significantly better performance. For example, in cross-scene testing (training on the Kitchen scene), Hg-I2P improves IR and RR by $\mathbf{7.9\%}$ and $\mathbf{11.9\%}$ over Top-I2P \cite{top-i2p}. Table \ref{tab:supp_exp_1} shows that Hg-I2P achieves more accurate RTE and RRE metrics. Hg-I2P's superiority arises from three key factors: (i) a heterogeneous graph that systematically encodes 2D-2D, 3D-3D, and 2D-3D relations, whereas Top-I2P does not; (ii) multi-path relation mining, rather than the trivial GCN for 2D-3D region matching \cite{top-i2p}; (iii) a complete cross-modal feature adaptation and refinement pipeline, compared to Top-I2P's simple interaction module. 

\begin{figure*}[t]
	\centering
		\includegraphics[width=1.0\linewidth]{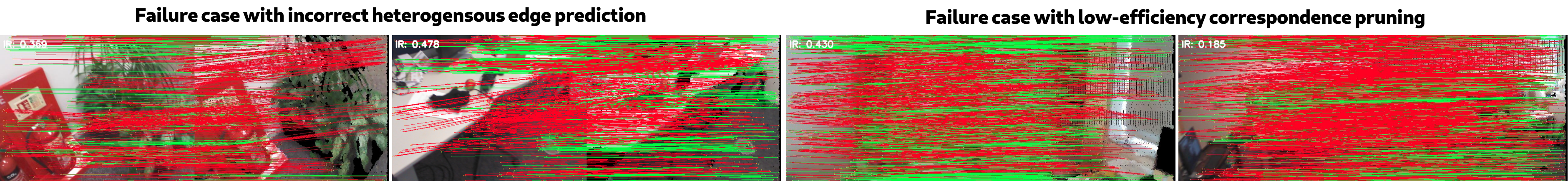}	
		\caption{Examples where Hg-I2P fails to generate reliable correspondences. Errors arise from incorrect heterogeneous edge prediction and overly relaxed pruning thresholds, often triggered by SAM over-segmentation or highly cluttered scenes. Green and red lines denote correct and incorrect correspondences, respectively.}
	\label{fig:supp_failure_heter}
\end{figure*}

MinCD \cite{mincd} enhances I2P registration by improving 2D-3D correspondence learning. However, in cross-scene and cross-dataset experiments, Hg-I2P consistently achieves higher IR and RR, indicating that a comprehensive cross-modal feature adaptation is more effective than refining only the correspondence loss. Overall, these results demonstrate that Hg-I2P achieves stronger generalization in indoor scenes. 

\vspace{+2mm}
\noindent\textbf{In-depth analysis on driving scenes}. Hg-I2P was initially designed for indoor scenes and cannot be applied directly to outdoor scenarios because the overlap between LiDAR point clouds and RGB images is limited. To address this, we used pre-trained CorrI2P \cite{corr-tcsvt} to extract the overlapped LiDAR region corresponding to the RGB image. Depth ranges also differ dramatically between indoor and outdoor scenes, making KPConv's fixed radius unsuitable. To avoid radius retuning, we scale the LiDAR point cloud such that its maximum depth is below $5.0m$, and incorporate this scaling in the RTE computation. 

Compared with outdoor-focused approaches such as CoFiI2P \cite{CoFiI2P}, CMR-Agent \cite{cmr-agent}, GraphI2P \cite{graph-i2p-cvpr}, and ImCorr \cite{corr-learning-i2p-cvpr}, Hg-I2P achieves more accurate performance because of : (i) fine-grained cross-modal feature interaction via the heterogeneous graph; (ii) pruning mechanism that reduces invalid correspondences; (iii) leveraging the overlapped point cloud to simplify I2P registration. 

Visualizations of Hg-I2P and MATR \cite{2d3d-matr} on the KITTI dataset \cite{kitti} are presented in Fig. \ref{fig:supp_kitti}. It shows that the fine-tuned Hg-I2P predicts the high-quality correspondences across diverse traffic scenes. We also provide the SAM results on the KITTI dataset \cite{kitti} in Fig. \ref{fig:supp_kitti_sam}, which shows that SAM performs reliably outdoors, contributing to the strong RTE and RRE metrics on the KITTI dataset \cite{kitti}. 

\section*{Appendix D. More Qualitative Comparisons}

Beyond the KITTI visualizations in Fig. \ref{fig:supp_kitti}, we provide additional qualitative results. Comparisons among Top-I2P \cite{top-i2p}, MinCD \cite{mincd}, and Hg-I2P are provided in Fig. \ref{fig:supp_qc_2}. We observe that Hg-I2P maintains stable performance across varied indoor scenes as it produces significantly fewer outliers.  We also evaluate HC-pruning, and the qualitative comparisons are shown in Fig. \ref{fig:supp_qc_1}, which further reduces outliers and improves IR.

\vspace{+2mm}
\noindent\textbf{Failure case analysis}. Despite outperforming state-of-the-art methods on many public datasets, Hg-I2P still fails in certain cases, as shown in Fig. \ref{fig:supp_failure_heter}. In complex scenes, SAM may produce over-segmentation, leading to overly complex graph connectivity and inaccurate heterogeneous graph edge prediction. These inaccuracies can distort feature adaptation and produce correspondences with a larger number of outliers. Additionally, the pruning criteria are sensitive to their thresholds: tight thresholds remove true inliers, while loose thresholds fail to remove outliers. Future work will explore improved heterogeneous edge prediction and correspondence pruning strategies. 

% Besides, to further investigate the effect of HE-adapting, we visualize the cross-modal features $\mathbf{F}_I$ and $\mathbf{F}_P$ (i.e., before HE-adapting) and $\mathbf{G}_I$ and $\mathbf{G}_P$ (i.e., after HE-adapting) where results are provided in Fig. 7. These features lie in a space with high dimensions, so that we project them in a 3D space with principal component analysis (PCA) for the better visualization. It indicates that HE-adapting enhances the cross-modal feature alignment and improves the performance of I2P registration. 

\section*{Appendix E. More Hyperparameter Experiments}

The main hyper-parameters in Hg-I2P are $\alpha$, $\delta_{\text{rej}}$, $\tau_{\text{rej}}$, and $\lambda_1$. We discuss their selections in this section. 

(1) $\alpha$ controls the adjacency structure of 2D-2D and 3D-3D regions. If $\alpha$ approaches zero, the adjacency matrices lose sparsity; if too large, they become overly sparse. Both extremes degrade heterogeneous edge prediction. We set $\alpha = 1.6$, and the performance is stable when $\alpha \in [0.5,2]$. 

\begin{table}[t]
%\scriptsize
\centering
\caption{Ablation results showing how the geometric reprojection threshold $\delta_{\text{rej}}$ influences registration recall. Moderate thresholds (e.g., $\delta_{\text{rej}} = 15$) achieve the best balance between retaining inliers and rejecting misaligned correspondences.}
\resizebox{1\linewidth}{!}{%}
\begin{tabular}{c|ccccc}
% \toprule
\hline
$\delta_{\text{rej}}$  & 5 & 10 & 15 & 20 & 25 \\
\hline
Registration recall (RR)    & 0.587 & 0.652 & \textcolor{blue}{\textbf{0.682}} & 0.670 & 0.662 \\
\hline
% \bottomrule
\end{tabular}}
\label{tab:supp_ab_1}
\end{table}

\begin{table}[t]
%\scriptsize
\centering
\caption{Comparison of fixed and top-K adaptive cosine-similarity threshold $\tau_{\text{rej}}$ in HC-pruning. Adaptive selection improves both inlier ratio and registration recall.}
\resizebox{1\linewidth}{!}{%}
\begin{tabular}{c|cc}
% \toprule
\hline
Scheme of $\tau_{\text{rej}}$  & Fixed  & Adaptive (Top-K selection)  \\
\hline
Inlier ratio (IR)           & 0.537 & \textcolor{blue}{\textbf{0.558}}  \\
Registration recall (RR)    & 0.685 & \textcolor{blue}{\textbf{0.690}}  \\
\hline
% \bottomrule
\end{tabular}}
\label{tab:supp_ab_2}
\end{table}

(2) $\delta_{\text{rej}}$ is the threshold in criterion I of HC-pruning. Because $\mathbf{\tilde{T}}$ is not perfectly accurate, a small threshold incorrectly removes many inliers (see Table \ref{tab:supp_ab_1}). Experiments show that $\delta_{\text{rej}} = 15$ yields the best RR. 

\begin{table}[t]
%\scriptsize
\centering
\caption{Ablation study on $\lambda_1$ (without HC-pruning), demonstrating that moderate weighting yields optimal registration recall by balancing heterogeneous edge prediction and correspondence learning.}
\resizebox{1\linewidth}{!}{%}
\begin{tabular}{c|ccccc}
% \toprule
\hline
$\lambda_1$     & 0.032 & 0.048 & 0.064 & 0.080 & 0.096 \\
\hline
Registration recall (RR)    & 0.602 & 0.638 & \textcolor{blue}{\textbf{0.661}} & 0.652 & 0.626 \\
\hline
% \bottomrule
\end{tabular}}
\label{tab:supp_ab_3}
\end{table}

(3) $\tau_{\text{rej}}$ is the threshold in criterion II for HC-pruning. We find that an adaptive, top-K strategy improves IR and RR. Specifically, we sort correspondences in descending order by cosine similarity and retain the top $85\%$. The ablation results are provided in Table \ref{tab:supp_ab_2}. 

(4) $\lambda_1$ controls the weight of heterogeneous edge prediction in the total loss. Because $L_{\mathrm{corr}}$ dominates cross-modal learning, $\lambda_1$ must remain below a certain value to ensure
\begin{equation}
\lambda_1 \leq L_{\mathrm{corr}}/\Vert \mathbf{\hat{E}}_{\text{I2P}}[\mathrm{mask}]-\mathbf{{E}}_{\text{I2P}}[\mathrm{mask}]\Vert_2^2
\end{equation}
Guided by the inequality in (A5), we empirically set $\lambda_1 \leq 0.1$. The ablation result of $\lambda_1$ is provided in Table \ref{tab:supp_ab_3}. If too small, heterogeneous edge prediction deteriorates; if too large, the model overemphasizes it at the expense of registration. The optimal value is set to $0.064$. 

\begin{table}[t]
\centering
\caption{Runtime of each pipeline component, including SAM segmentation, graph initialization (including feature extraction), multi-path relation mining, heterogeneous edge adaptation, and HC-pruning. Graph initialization and SAM dominate the total runtime (unit: ms).}
\resizebox{1.0\linewidth}{!}{%}
\begin{tabular}{c|ccccc|c}
% \toprule
\hline
Proc. & SAM & Ini. & MP-mining & HE-adapting & HC-pruning & All \\
\hline
Time & 47.2 & 72.4 & 3.9 & 34.6 & 19.7 & 177.8   \\
\hline
%\bottomrule
\end{tabular}}
\label{tab:ab_4}
\end{table}

\begin{figure}[t]
	\centering
		\includegraphics[width=1.0\linewidth]{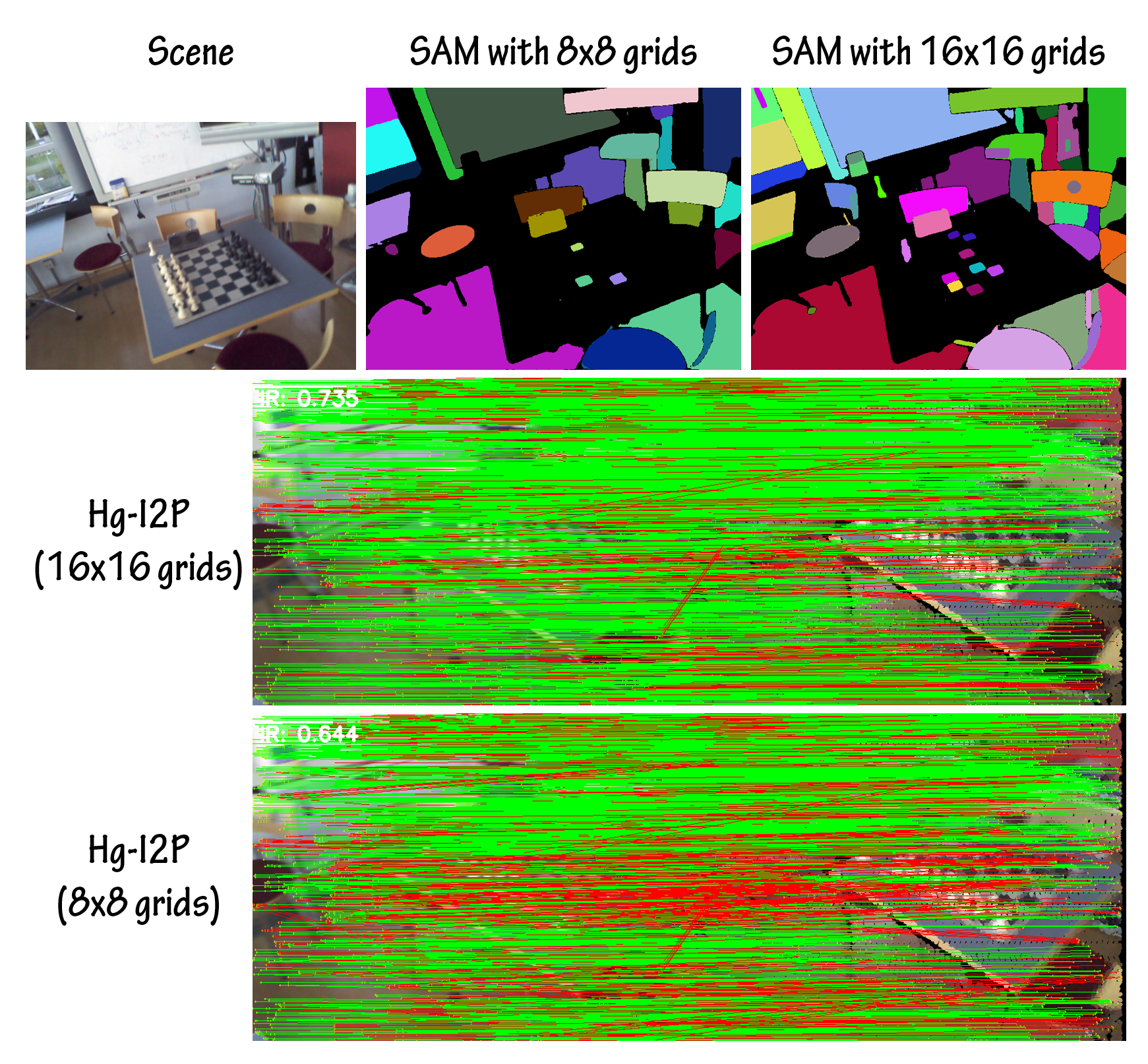}	
		\caption{Visualization of how different grid-based prompting schemes for SAM (e.g., $8\times 8$ vs. $16\times 16$) affect segmentation granularity and the resulting 2D–3D correspondences. Denser prompting generally improves alignment but increases inference time. Green and red lines denote correct and incorrect correspondences, respectively.}
	\label{fig:supp_future}
\end{figure}

\begin{table}[t]
%\scriptsize
\centering
\caption{Evaluation of segmentation prompt density ($8\times 8$ vs. $16\times 16$ grid points) on registration accuracy and inference time. Denser prompting provides higher registration recall at the cost of additional computation.}
\resizebox{1.0\linewidth}{!}{%}
\begin{tabular}{c|cc}
% \toprule
\hline
Prompt of 2D SAM  & RR & Time  \\
\hline
$8\times 8$ grid points  & 62.1 & \textcolor{blue}{\textbf{144.2 ms}} \\
$16\times 16$ grid points (used in experiments) & \textcolor{blue}{\textbf{68.2}} & 177.8 ms \\
\hline
% \bottomrule
\end{tabular}}
\label{tab:supp_ab_prompt}
\end{table}

% \vspace{+2mm}
% \noindent\textbf{Runtime analysis.} 
We also evaluate the runtime of Hg-I2P, and the result is provided in Table \ref{tab:ab_4}. Hg-I2P runs at 5.6 FPS, compared to 12.8 FPS for MATR \cite{2d3d-matr}. Most time is spent on 2D SAM and graph initialization. Although slower, Hg-I2P offers stronger generalization.  

Finally, we analyze the effect of 2D SAM prompting. From Fig. \ref{fig:supp_future}, it is observed that the segmentation number is different with prompts. Using denser grid prompts (e.g., $16\times 16$ grid points) produces more regions and improves registration performance, but increases inference time (see Table \ref{tab:supp_ab_prompt}). Applications may choose prompts based on accuracy and speed requirements.

\section*{Appendix F. Limitations and Future Works}

Hg-I2P primarily depends on pre-trained 2D and 3D SAM. In complex scenes, it may produce over- or under-segmentation, especially for outdoor scenes. Over-segmentation complicates heterogeneous edge prediction and reduces registration stability, as shown in Fig. \ref{fig:supp_failure_heter}. Moreover, the inference time for 2D and 3D SAM is substantially higher than that of the other pipeline components. 
To address these limitations, we plan to explore constructing and applying heterogeneous graphs based on conformal geometry, which may offer a more reliable structure for cross-modal representation.

{
    \small
    \bibliographystyle{ieeenat_fullname}
    \bibliography{main}

@String(AAAI = {AAAI})

@article{camera_model,
  author       = {Zhengyou Zhang},
  title        = {A Flexible New Technique for Camera Calibration},
  journal      = {{IEEE} Trans. Pattern Anal. Mach. Intell.},
  volume       = {22},
  number       = {11},
  pages        = {1330--1334},
  year         = {2000},
}

@inproceedings{top-i2p,
  author       = {Pei An and Jiaqi Yang and Muyao Peng and et al.},
  title        = {{Top-I2P}: Explore Open-Domain Image-to-Point Cloud Registration Using Topology Relationship},
  booktitle    = {Proceedings of International Joint Conference on
                  Artificial Intelligence},
  pages        = {1--9},
  year         = {2025},
}

@inproceedings{kpconv,
  author       = {Hugues Thomas and
                  Charles R. Qi and
                  Jean{-}Emmanuel Deschaud and
                  et al.},
  title        = {KPConv: Flexible and Deformable Convolution for Point Clouds},
  booktitle    = {Proceedings of {IEEE/CVF} International Conference on Computer Vision},
  pages        = {6410--6419},
  year         = {2019},
}

@inproceedings{resnet,
  author       = {Kaiming He and Xiangyu Zhang and Shaoqing Ren and Jian Sun},
  title        = {Deep Residual Learning for Image Recognition},
  booktitle    = {Proceedings of {IEEE} Conference on Computer Vision and Pattern Recognition},
  pages        = {770--778},
  year         = {2016},
}

@inproceedings{sam_2d,
  author={Alexander Kirillov and
  Eric Mintun and
  Nikhila Ravi and et al.},
  booktitle={Proceedings of {IEEE/CVF} International Conference on Computer Vision},
  title={Segment Anything},
  year={2023},
  pages={3992--4003},
}

@inproceedings{sam_3d,
  author={Yuchen Zhou and
                  Jiayuan Gu and
                  Tung Yen Chiang and
                  et al.},
  booktitle={Proceedings of International Conference on Learning Representations},
  title={Point-SAM: Promptable 3D Segmentation Model for Point Clouds},
  year={2025},
  pages={1--10},
}

@inproceedings{transformer,
  author       = {Ashish Vaswani and
                  Noam Shazeer and
                  Niki Parmar and
                  et al.},
  title        = {Attention is All you Need},
  booktitle    = {Proceedings of Advances in Neural Information Processing Systems},
  pages        = {5998--6008},
  year         = {2017},
}

@inproceedings{2d3d-matr,
  author       = {Minhao Li and
                  Zheng Qin and
                  Zhirui Gao and
                  et al.},
  title        = {{2D3D-MATR}: {2D-3D} Matching Transformer for Detection-free Registration
                  between Images and Point Clouds},
  booktitle    = {Proceedings of {IEEE} Conference on Computer Vision},
  pages        = {1--10},
  year         = {2023},
}

@inproceedings{bridge,
  author       = {Zhixin Cheng and
                  Jiacheng Deng and
                  Xinjun Li and
                  et al.},
  title        = {Bridge 2D-3D: Uncertainty-aware Hierarchical Registration Network
                  with Domain Alignment},
  booktitle    = {Proceedings of {AAAI} Conference on Artificial Intelligence},
  pages        = {2491--2499},
  year         = {2025},
}

@inproceedings{diff-reg,
  author       = {Qianliang Wu and
                  Haobo Jiang and
                  Lei Luo and
                  et al.},
  title        = {Diff-Reg: Diffusion Model in Doubly Stochastic Matrix Space for Registration
                  Problem},
  booktitle    = {Proceedings of {IEEE/CVF} European Conference on Computer Vision},
  volume       = {15123},
  pages        = {160--178},
  year         = {2024},
}

@article{epnp,
  author       = {Vincent Lepetit and
                  Francesc Moreno{-}Noguer and
                  Pascal Fua},
  title        = {{EP\emph{n}P}: An Accurate \emph{O}(\emph{n}) Solution to the P\emph{n}P
                  Problem},
  journal      = {Int. J. Comput. Vis.},
  volume       = {81},
  number       = {2},
  pages        = {155--166},
  year         = {2009},
}

@inproceedings{circle-loss,
  author       = {Yifan Sun and
                  Changmao Cheng and
                  Yuhan Zhang and
                  et al.},
  title        = {Circle Loss: {A} Unified Perspective of Pair Similarity Optimization},
  booktitle    = {Proceedings of {IEEE/CVF} Conference on Computer Vision and Pattern Recognition},
  pages        = {6397--6406},
  year         = {2020},
}

@article{flowi2p,
  author       = {Pei An and
                  You Yang and
                  Jiaqi Yang and
                  Muyao Peng and
                  Qiong Liu and
                  Liangliang Nan},
  title        = {Enhance Image-to-Point-Cloud Registration with Beltrami Flow},
  journal      = {Int. J. Comput. Vis.},
  volume       = {133},
  number       = {12},
  pages        = {8589--8616},
  year         = {2025},
}

@inproceedings{furao,
  author       = {Rao Fu and
                  Jianmin Zheng and
                  Liang Yu},
  title        = {Consistent Normal Orientation for 3D Point Clouds via Least Squares on Delaunay Graph},
  booktitle    = {Proceedings of {IEEE/CVF} Conference on Computer Vision and Pattern Recognition},
  pages        = {16932--16942},
  year         = {2025},
}

@inproceedings{2d3d-match,
  author       = {Mengdan Feng and
                  Sixing Hu and
                  Marcelo H. Ang and
                  Gim Hee Lee},
  title        = {{2D3D-Matchnet}: Learning To Match Keypoints Across 2D Image And 3D
                  Point Cloud},
  booktitle    = {Proceedings of {IEEE} International Conference on Robotics and Automation},
  pages        = {4790--4796},
  year         = {2019},
}

@inproceedings{ep2p-loc,
  author       = {Minjung Kim and
                  Junseo Koo and
                  Gunhee Kim},
  title        = {EP2P-Loc: End-to-End 3D Point to 2D Pixel Localization for Large-Scale
                  Visual Localization},
  booktitle    = {Proceedings of {IEEE/CVF} International Conference on Computer Vision},
  pages        = {21470--21480},
  year         = {2023},
}

@inproceedings{iplannar,
  author       = {Fan Yang and
                  Chen Wang and
                  Cesar Cadena and
                  Marco Hutter},
  title        = {iPlanner: Imperative Path Planning},
  booktitle    = {Proceedings of Robotics: Science and Systems},
  pages        = {1--9},
  year         = {2023},
}

@inproceedings{mast3r-slam,
  author       = {Riku Murai and
                  Eric Dexheimer and
                  Andrew J. Davison},
  title        = {MASt3R-SLAM: Real-Time Dense SLAM with 3D Reconstruction Priors},
  booktitle    = {Proceedings of {IEEE/CVF} Conference on Computer Vision and Pattern Recognition},
  pages        = {1--10},
  year         = {2025},
}

@inproceedings{deep-i2p,
  author       = {Jiaxin Li and
                  Gim Hee Lee},
  title        = {{DeepI2P}: Image-to-Point Cloud Registration via Deep Classification},
  booktitle    = {Proceedings of {IEEE} Conference on Computer Vision and Pattern Recognition},
  pages        = {15960--15969},
  year         = {2021},
}

@article{corr-tcsvt,
  author       = {Siyu Ren and
                  Yiming Zeng and
                  Junhui Hou and
                  Xiaodong Chen},
  title        = {{CorrI2P}: Deep Image-to-Point Cloud Registration via Dense Correspondence},
  journal      = {{IEEE} Trans. Circuits Syst. Video Technol.},
  volume       = {33},
  number       = {3},
  pages        = {1198--1208},
  year         = {2023},
}

@inproceedings{freereg,
  author       = {Haiping Wang and
                  Yuan Liu and
                  Bing Wang and
                  et al.},
  title        = {FreeReg: Image-to-Point Cloud Registration Leveraging Pretrained Diffusion Models and Monocular Depth Estimators},
  booktitle    = {Proceedings of International Conference on Learning Representation},
  pages        = {1--24},
  year         = {2024},
}

@inproceedings{p2-net,
  author       = {Bing Wang and
                  Changhao Chen and
                  Zhaopeng Cui and
                  et al.},
  title        = {{P2-Net}: Joint Description and Detection of Local Features for Pixel
                  and Point Matching},
  booktitle    = {Proceedings of {IEEE} International Conference on Computer Vision},
  pages        = {15984--15993},
  year         = {2021},
}

@inproceedings{graph-i2p-cvpr,
  author       = {Lin Bie and Shouan Pan and Siqi Li and et al.},
  title        = {GraphI2P: Image-to-Point Cloud Registration with Exploring Pattern of Correspondence via Graph Learning},
  booktitle    = {Proceedings of {IEEE} Conference on Computer Vision and Pattern Recognition},
  pages        = {22161--22171},
  year         = {2025},
}

@inproceedings{corr-learning-i2p-cvpr,
  author       = {Xinjun Li and Wenfei Yang and Jiacheng Deng and et al.},
  title        = {Implicit Correspondence Learning for Image-to-Point Cloud Registration},
  booktitle    = {Proceedings of {IEEE} Conference on Computer Vision and Pattern Recognition},
  pages        = {16922--16931},
  year         = {2025},
}

@inproceedings{RGBD-dataset,
  author       = {Kevin Lai and
                  Liefeng Bo and
                  Dieter Fox},
  title        = {Unsupervised feature learning for 3D scene labeling},
  booktitle    = {Proceedings of {IEEE} International Conference on Robotics and Automation},
  pages        = {3050--3057},
  year         = {2014},
}

@article{CoFiI2P,
  author       = {Shuhao Kang and
                  Youqi Liao and
                  Jianping Li and
                  et al.},
  title        = {CoFiI2P: Coarse-to-Fine Correspondences-Based Image to Point Cloud
                  Registration},
  journal      = {{IEEE} Robotics Autom. Lett.},
  volume       = {9},
  number       = {11},
  pages        = {10264--10271},
  year         = {2024},
}

@inproceedings{cmr-agent,
  author       = {Gongxin Yao and
                  Yixin Xuan and
                  Xinyang Li and
                  Yu Pan},
  title        = {CMR-Agent: Learning a Cross-Modal Agent for Iterative Image-to-Point
                  Cloud Registration},
  booktitle    = {Proceedings of {IEEE/RSJ} International Conference on Intelligent Robots and Systems},
  pages        = {13458--13465},
  year         = {2024},
}

@inproceedings{scene-7-dataset,
  author       = {Ben Glocker and
                  Shahram Izadi and
                  Jamie Shotton and
                  Antonio Criminisi},
  title        = {Real-time {RGB-D} camera relocalization},
  booktitle    = {Proceedings of {IEEE} International Symposium on Mixed and Augmented Reality},
  pages        = {173--179},
  year         = {2013},
}

@inproceedings{scan-net,
  author       = {Angela Dai and
                  Angel X. Chang and
                  Manolis Savva and
                  et al.},
  title        = {ScanNet: Richly-Annotated 3D Reconstructions of Indoor Scenes},
  booktitle    = {Proceedings of {IEEE} Conference on Computer Vision and Pattern Recognition},
  pages        = {2432--2443},
  year         = {2017},
}

@inproceedings{kitti,
  author       = {Andreas Geiger and
                  Philip Lenz and
                  Raquel Urtasun},
  title        = {Are we ready for autonomous driving? The {KITTI} vision benchmark
                  suite},
  booktitle    = {Proceedings of {IEEE} Conference on Computer Vision and Pattern Recognition},
  pages        = {3354--3361},
  year         = {2012},
}

@inproceedings{tum,
  author       = {J{\"{u}}rgen Sturm and
                  Nikolas Engelhard and
                  Felix Endres and
                  et al.},
  title        = {A benchmark for the evaluation of {RGB-D} {SLAM} systems},
  booktitle    = {Proceedings of {IEEE/RSJ} International Conference on Intelligent Robots and
                  Systems},
  pages        = {573--580},
  year         = {2012},
}

@inproceedings{diff_reg_match,
  author       = {Junsheng Zhou and
                  Baorui Ma and
                  Wenyuan Zhang and
                  et al.},
  title        = {Differentiable Registration of Images and LiDAR Point Clouds with
                  VoxelPoint-to-Pixel Matching},
  booktitle    = {Proceedings of Advances in Neural Information Processing Systems},
  pages        = {1--10},
  year         = {2023},
}

@inproceedings{diff-pnp-layer,
  author       = {Hansheng Chen and
                  Pichao Wang and
                  Fan Wang and
                  et al.},
  title        = {EPro-PnP: Generalized End-to-End Probabilistic Perspective-n-Points
                  for Monocular Object Pose Estimation},
  booktitle    = {Proceedings of {IEEE/CVF} Conference on Computer Vision and Pattern Recognition},
  pages        = {2771--2780},
  year         = {2022},
}

@article{ol-reg,
  author       = {Pei An and
                  Xuzhong Hu and
                  Junfeng Ding and
                  et al.},
  title        = {OL-Reg: Registration of Image and Sparse LiDAR Point Cloud with Object-Level Dense Correspondences},
  journal      = {{IEEE} Trans. Circuits Syst. Video Technol.},
  volume       = {34},
  number       = {8},
  pages        = {7523--7536},
  year         = {2024},
}

@inproceedings{geotrans,
  author       = {Zheng Qin and
                  Hao Yu and
                  Changjian Wang and
                  et al.},
  title        = {Geometric Transformer for Fast and Robust Point Cloud Registration},
  booktitle    = {Proceedings of {IEEE/CVF} Conference on Computer Vision and Pattern Recognition},
  pages        = {11133--11142},
  year         = {2022},
}

@article{PAPI-Reg,
  author       = {Yuanchao Yue and
                  Zhengxin Li and
                  Wei Zhang and
                  Hui Yuan},
  title        = {PAPI-Reg: Patch-to-Pixel Solution for Efficient Cross-Modal Registration
                  between LiDAR Point Cloud and Camera Image},
  journal      = {CoRR},
  volume       = {abs/2503.15285},
  year         = {2025},
}

@inproceedings{depthanything,
  author       = {Lihe Yang and
                  Bingyi Kang and
                  Zilong Huang and
                  et al.},
  title        = {Depth Anything: Unleashing the Power of Large-Scale Unlabeled Data},
  booktitle    = {Proceedings of {IEEE/CVF} Conference on Computer Vision and Pattern Recognition},
  pages        = {10371--10381},
  year         = {2024},
}

@inproceedings{controlnet,
  author       = {Lvmin Zhang and
                  Anyi Rao and
                  Maneesh Agrawala},
  title        = {Adding Conditional Control to Text-to-Image Diffusion Models},
  booktitle    = {Proceedings of {IEEE/CVF} International Conference on Computer Vision},
  pages        = {3813--3824},
  year         = {2023},
}

@inproceedings{dino,
  author       = {Mathilde Caron and
                  Hugo Touvron and
                  Ishan Misra and
                  et al.},
  title        = {Emerging Properties in Self-Supervised Vision Transformers},
  booktitle    = {Proceedings of {IEEE/CVF} International Conference on Computer Vision},
  pages        = {9630--9640},
  year         = {2021},
}

@inproceedings{mast3r,
  author       = {Vincent Leroy and
                  Yohann Cabon and
                  J{\'{e}}r{\^{o}}me Revaud},
  title        = {Grounding Image Matching in 3D with MASt3R},
  booktitle    = {Proceedings of European Conference on Computer Vision},
  pages        = {71--91},
  year         = {2024},
}

@inproceedings{centerpts,
  author       = {Tianwei Yin and
                  Xingyi Zhou and
                  Philipp Kr{\"{a}}henb{\"{u}}hl},
  title        = {Center-Based 3D Object Detection and Tracking},
  booktitle    = {Proceedings of {IEEE} Conference on Computer Vision and Pattern Recognition},
  pages        = {11784--11793},
  year         = {2021},
}

@inproceedings{sam3d_used,
  author={Yunhan Yang and Xiaoyang Wu and Tong He and et al.},
  title        = {SAM3D: Segment Anything in 3D Scenes},
  booktitle    = {Proceedings of {IEEE/CVF} International Conference on Computer Vision Workshops},
  pages        = {1--5},
  year         = {2023},
}

@article{fastsam,
  author       = {Xu Zhao and
                  Wenchao Ding and
                  Yongqi An and et al.},
  title        = {Fast Segment Anything},
  journal      = {CoRR},
  volume       = {abs/2306.12156},
  year         = {2023},
  pages        = {1--11},
}

@inproceedings{lcd,
  author       = {Quang{-}Hieu Pham and
                  Mikaela Angelina Uy and
                  Binh{-}Son Hua and
                  et al.},
  title        = {{LCD:} Learned Cross-Domain Descriptors for 2D-3D Matching},
  booktitle    = {Proceedings of {AAAI} Conference on Artificial Intelligence},
  pages        = {11856--11864},
  year         = {2020},
}

@inproceedings{superglue,
  author       = {Paul{-}Edouard Sarlin and
                  Daniel DeTone and
                  Tomasz Malisiewicz and
                  Andrew Rabinovich},
  title        = {SuperGlue: Learning Feature Matching With Graph Neural Networks},
  booktitle    = {Proceedings of {IEEE/CVF} Conference on Computer Vision and Pattern Recognition},
  pages        = {4937--4946},
  year         = {2020},
}

@inproceedings{diffi2p,
  author       = {Juncheng Mu and
                  Changwei Ren and
                  Weixiang Zhang and et al.},
  title        = {Diff$^2$I2P: Differentiable Image-to-Point Cloud registration with Diffusion Prior},
  booktitle    = {Proceedings of {IEEE/CVF} International Conference on Computer Vision},
  pages        = {1--11},
  year         = {2025},
}

@inproceedings{mincd,
  author       = {Pei An and Jiaqi Yang and Muyao Peng and You Yang and Qiong Liu and Xiaolin Wu and Liangliang Nan},
  title        = {MinCD-PnP: Learning 2D-3D Correspondences with Approximate Blind PnP},
  booktitle    = {Proceedings of {IEEE/CVF} International Conference on Computer Vision},
  pages        = {1--11},
  year         = {2025},
}

@article{cmr-next,
  author       = {Daniele Cattaneo and
                  Abhinav Valada},
  title        = {CMRNext: Camera to LiDAR Matching in the Wild for Localization and
                  Extrinsic Calibration},
  journal      = {{IEEE} Trans. Robotics},
  volume       = {41},
  pages        = {1995--2013},
  year         = {2025},
}
}


\end{document}